%% file: main.tex
\documentclass{article}
\usepackage{iclr2027_conference,times}
\usepackage[T1]{fontenc}
\usepackage{amsmath,amssymb,booktabs,graphicx,url,array,longtable,float,multirow,xcolor,hyperref}
\title{RolloutFaith: Auditing Persistent Internal Interventions in Visual World Models}
\author{Junchi Yao\textsuperscript{*}\\
MBZUAI \And
Ziyi Wang\textsuperscript{*}\\
UESTC \And
Youling Huang\\
Dalian University of Technology \And
Lijie Hu\textsuperscript{\ensuremath{\dagger}}\\
MBZUAI
}
\newcommand{\isg}{\mathrm{ISG}}
\newcommand{\ssg}{\mathrm{SSG}}
\definecolor{improvedblue}{HTML}{E8F3FA}
\definecolor{improvementgreen}{HTML}{168247}
\definecolor{appendixblue}{HTML}{2F6F9F}
\definecolor{appendixpale}{HTML}{EDF5FA}
\definecolor{appendixslate}{HTML}{475569}
\newcommand{\improvedcell}[1]{{\setlength{\fboxsep}{1.2pt}\colorbox{improvedblue}{\strut #1}}}
\newcommand{\improvementarrow}{\textcolor{improvementgreen}{\ensuremath{\uparrow}}}
\iclrfinalcopy
\begin{document}
\maketitle
\begin{abstract}
Interpretability methods such as probes, activation patches and learned editors are designed to reveal or modify a model's current computation. World models pose a harder requirement: because their predictions become inputs to later predictions, a useful internal correction must survive after editing stops. We therefore propose \textbf{RolloutFaith}, a framework that measures semantic improvement both in the prediction produced at intervention time and over later autonomous predictions under fixed events, actions, noise, and information budgets. We evaluate ten fitted editors on three world models across Crafter, Cartpole, and CoinRun. We also use Reference Activation Patching, which replaces a model activation with the paired activation computed from the real observation, to measure the correction available at the chosen interface. This reference intervention improves later predictions in all nine model and task combinations and outperforms the best fitted editor in eight, yet its sustained gain decreases with horizon in five of nine combinations. Current fitted editors recover only limited and inconsistent long term effects. By restoring individual state components to their untouched values, we find that persistent effects travel through the newest generated frame in DIAMOND, recurrent memory in DreamerV3, and both in STORM. These findings suggest that training should reward future consequences. To test this hypothesis, we propose \emph{Delayed LoReFT}, which optimizes the same low rank intervention through four frozen future transitions and improves sustained intervention effects to some extent.
\end{abstract}

\input{figures/teaser.tex}
\input{sections/01_introduction.tex}
\input{sections/02_related_work.tex}
\input{sections/03_framework_and_metrics.tex}
\input{sections/04_experimental_results.tex}
\input{sections/05_component_analysis.tex}
\input{sections/06_rolloutfaith_improvement.tex}
\input{sections/07_conclusion.tex}
\phantomsection\label{end:maintext}
\clearpage

\section*{Reproducibility Statement}\label{statement:repro}
The event, information and measurement contracts are specified in Section~\ref{sec:experiments}. Appendix~\ref{app:implementation} gives the complete setup and fitting details. Appendices~\ref{app:statistics} through~\ref{app:temporal} report uncertainty, transfer, matched controls and temporal robustness. Appendices~\ref{app:diagnostics} through~\ref{app:local-propagation} document metric validation, component pathways and the limits of local interpretation. All fitted editors use development-only fitting and selection. The diagnostic extensions reuse the same checkpoints, editors, events, actions, and noise seeds. Audits of saved frames verify branch identities, realized norms, component restoration, and exact recovery of the untouched future under joint restoration. All aggregates use equal source weights after averaging registered seeds and events within source. Figure~\ref{fig:teaser} uses an audited event from the formal CoinRun cohort.

\section*{Artificial Intelligence (AI) Use Statement}\label{statement:ai}
We used ChatGPT as an assistant to refine the manuscript, improve clarity, and enhance the structure and readability.

\bibliography{references}
\bibliographystyle{iclr2027_conference}

\clearpage

\appendix
\setcounter{topnumber}{4}
\setcounter{bottomnumber}{2}
\setcounter{totalnumber}{6}
\renewcommand{\topfraction}{0.95}
\renewcommand{\bottomfraction}{0.85}
\renewcommand{\textfraction}{0.05}
\renewcommand{\floatpagefraction}{0.75}
\phantomsection
\addcontentsline{toc}{section}{Appendix Contents}
\begingroup
\begin{center}
{\fontsize{18}{21}\selectfont\bfseries\textcolor{appendixblue}{Appendix Contents}}\par
\vspace{4pt}
\textcolor{appendixblue}{\rule{0.28\linewidth}{1.1pt}}
\end{center}
\vspace{3pt}

\setlength{\parskip}{0pt}
\newcommand{\appsectionentry}[2]{%
  \vspace{3.5pt}%
  \noindent
  {\setlength{\fboxsep}{4.5pt}%
  \colorbox{appendixpale}{%
    \parbox{\dimexpr\linewidth-2\fboxsep\relax}{%
      \hyperref[#1]{\textcolor{appendixblue}{\large\bfseries Appendix~\ref*{#1}\enspace #2}}%
      \hfill\textcolor{appendixblue}{\large\bfseries\pageref{#1}}%
    }%
  }}%
  \par\vspace{1.8pt}%
}
\newcommand{\appsubsectionentry}[2]{%
  \noindent\hspace*{1.35em}%
  \textcolor{appendixblue}{\raisebox{0.35ex}{\rule{0.42em}{0.42em}}}%
  \hspace{0.55em}{\normalsize\hyperref[#1]{\textcolor{appendixslate}{\ref*{#1}\enspace #2}}%
  \dotfill\textcolor{appendixblue}{\pageref{#1}}}\par\vspace{0.7pt}%
}
\appsectionentry{app:implementation}{Experimental Setup and Reproducibility}
\appsubsectionentry{app:setup-scope}{Evaluation sources and task qualification}
\appsubsectionentry{app:interfaces-features}{Intervention interfaces and semantic features}
\appsubsectionentry{app:editor-fitting}{Editor implementations, fitting and selection}
\appsubsectionentry{app:wm-training}{World-model training and checkpoints}
\appsubsectionentry{app:diagnostic-protocol}{Additional diagnostic protocol}
\appsectionentry{app:statistics}{Complete Intervention Results and Uncertainty}
\appsubsectionentry{app:modelwise-results}{Model-wise intervention profiles}
\appsubsectionentry{app:coverage-retention}{Scoreability and endpoint retention}
\appsubsectionentry{app:special-cases}{Interface headroom and the DreamerV3 exception}
\appsectionentry{app:transfer-controls}{Transfer and Random Controls}
\appsubsectionentry{app:matched-random-controls}{Matched random directions and requests with zero change}
\appsubsectionentry{app:complete-control-matrix}{Complete validation, test and random control matrix}
\appsectionentry{app:magnitude-direction}{Magnitude and Direction Controls}
\appsubsectionentry{app:reference-at-fitted-magnitude}{Reference direction at fitted magnitude}
\appsubsectionentry{app:fitted-at-reference-magnitude}{Fitted direction at reference magnitude}
\appsubsectionentry{app:reference-timing}{Reference timing}
\appsectionentry{app:temporal}{Temporal Robustness and Long Horizon Behavior}
\appsubsectionentry{app:temporal-sensitivity}{Sensitivity to horizon, action and prediction noise}
\appsubsectionentry{app:long-horizon-regimes}{Representative long-horizon regimes}
\appsectionentry{app:diagnostics}{Metric Validation and Qualitative Examples}
\appsubsectionentry{app:measurement-competence}{Prediction and measurement competence}
\appsubsectionentry{app:task-examples}{Task overview examples}
\appsubsectionentry{app:fixed-native-example}{Fixed native example}
\appsectionentry{app:temporal-pathways}{Component Restoration and Pathway Analysis}
\appsubsectionentry{app:pathway-protocol}{Protocol and complete aggregate results}
\appsubsectionentry{app:pathway-time-resolved}{Time resolved component pathways}
\appsubsectionentry{app:pathway-hybrid-controls}{Interpretation and controls with hybrid states}
\appsectionentry{app:local-propagation}{Local Analysis and Interpretation Limits}
\appsubsectionentry{app:local-conditions}{Local propagation conditions}
\appsubsectionentry{app:direction-ranking}{Direction, magnitude and ranking reversal}
\appsubsectionentry{app:restoration-limits}{State restoration and interpretation limits}
\endgroup

\clearpage
\section{Experimental Setup and Reproducibility}\label{app:implementation}
\subsection{Evaluation sources and task qualification}\label{app:setup-scope}
\textbf{Scope and input contracts.}
All intervention methods use a common test roster, permitted observations and native edit boundary within each model and task. Fitting and validation sources are disjoint from each other and from confirmation sources. At evaluation, future real frames are available only to the offline scorer. Each task uses three fixed prediction seeds and the same selected events across models. Evaluation episodes are disjoint from all editor fitting and validation sources.

\textbf{Source distribution.}
Crafter and Cartpole each request 48 fresh episodes of at most 512 actions, using the frozen DreamerV3 actor specific to each task in evaluation mode. Candidate prediction roots are 96, 224 and 352; the first two eligible events per source are retained. CoinRun uses 48 new levels, indexed from 30000 to 30047, with 512 actions per level; resets remain within that level. Its behavior holds a sampled action for eight steps: left, no action, right and right with jump have probabilities 0.2, 0.2, 0.3 and 0.3. The earliest two eligible roots at or after step 96 are selected, separated by at least 40 steps, with a history of 64 frames without resets and a continuation of 32 frames. Eligibility depends on scoreability using real references. All WMs receive the resulting common task roster.

\textbf{Task qualification and excluded Breakout candidate.}
We initially considered Atari Breakout with ball localization as the observable. Native evaluation completed for all three WMs on the same 47-source, 86-event roster. Over the eight-step scoring horizon, however, the ball detector found the ball in 1,721 of 1,992 DIAMOND prediction frames (86.4\%), none of 1,992 DreamerV3 frames, and 101 of 1,992 STORM frames (5.1\%); the available STORM checkpoint also came from an interrupted training run. Breakout therefore failed the task-level qualification requirement that the observable remain measurable for every WM and was replaced by CoinRun, which passed the frozen numerical and generated-image checks for all three models before editor development. This exclusion reflects inadequate prediction and measurement coverage for a common comparison, rather than failure to execute either model or a claim of general architectural failure.

\subsection{Intervention interfaces and semantic features}\label{app:interfaces-features}
\textbf{Common comparison interfaces and inputs.}
DIAMOND replaces the full $64\times64\times64$ activation of the final U Net up block at the three denoiser calls of one frame. Its reference prediction uses four real frames ending at $t-1$, where $t$ denotes the edited prediction. DreamerV3 replaces a $32\times16$ array of categorical prior logits (32 stochastic variables with 16 categories each, 512 coordinates) before native unimix and Gumbel sampling. STORM replaces a $32\times32$ array of categorical prior logits (32 variables with 32 categories each, 1,024 coordinates) before sampling; singleton batch and time axes are omitted from these dimensions. Both use posterior references that include real image $t$. DreamerV3 preserves the recipient deterministic state; STORM preserves its transformer cache. All methods are permitted observations through $t$. The mean tool uses a fixed training delta, probes use the recipient and semantic target, and ridge, carrier, LoReFT, the nonlinear MLP and Concept DAS use recipient and request features. Manifold uses the recipient and current semantic target to select a training anchor. Only RAP directly inserts reference activations during evaluation. Each fitted editor is capped using the training distribution of correction norms; full RAP retains its native magnitude. Realized norms and matched random effects accompany every fitted-editor comparison.

\textbf{Feature and solver specifications.}
Cartpole, Crafter and CoinRun semantic descriptors have 6, 1,197 and 128 coordinates; concatenating target, current prediction, their residual and the action encoding gives request dimensions 19, 3,608 and 399. The Cartpole descriptor is extracted only from the rendered $64\times64$ image, without simulator state. For a visible cart--pole foreground mask, it is
\begin{equation}
 q=\left(x_c/64,\ y_c/64,\ \cos 2\theta,\ \sin 2\theta,\ \log(1+n)/8,\ 1\right),
\end{equation}
where $(x_c,y_c)$ is the foreground-pixel centroid, $\theta\in[0,\pi)$ is the principal-axis angle, and $n$ is the foreground-pixel count; masks with fewer than three pixels receive the zero vector. The doubled-angle coordinates encode an unoriented principal axis continuously. The Crafter descriptor is a $7\times9$ grid with a calibrated 19-way one-hot tile label per cell ($7\times9\times19=1{,}197$ coordinates). CoinRun uses an $8\times8$ grid with two terrain-class fractions per cell ($8\times8\times2=128$ coordinates). PCA plus ridge \citep{pearson1901lines,hoerl1970ridge} and the uncentered carrier retain rank eight. Inputs $[a;r]$ are standardized with training means and standard deviations floored at $10^{-3}$, and coefficient regression uses penalty $10^{-3}$. Probe readouts use the same penalty; pseudoinverses use cutoff $10^{-5}$. LoReFT searches ranks four and eight. DreamerV3 and STORM use categorical sampling with hard values in the forward pass and soft values in the backward pass, following the Gumbel--Softmax and straight-through gradient estimators \citep{jang2017categoricalreparameterization,bengio2013estimating}; DIAMOND retains its native quantized Euler forward and uses a clamp with direct gradients in training. The code for semantic scores is never differentiated through.

\subsection{Editor implementations, fitting and selection}\label{app:editor-fitting}
\textbf{Fitting and selection budgets.}
Each linear family uses a norm cap at the 90th percentile of training correction norms. Its dose grid is $\{0.01,0.03,0.1,0.25,0.5,1\}$. Selection maximizes validation SSG@8 averaged by source among candidates with measured nonzero edits. Harmful test effects remain in the comparison.

Each LoReFT objective searches ranks $\{4,8\}$ at learning rates $10^{-4}$ and $3\times10^{-5}$, for 256 updates per rank and rate, with a cap equal to one tenth of the cap for training correction norms. Native validation occurs at updates 0, 1, 4, 8, 16, 32, 64, 128 and 256; active selection excludes the initial identity. The earlier configuration, fitted with learning rate $10^{-3}$ and the full cap for 128 updates per rank, is retained as a candidate. A final paired replay compares the earlier configuration and two refined editors on identical development events and selects the highest SSG@8 averaged by source among active candidates. All 18 combinations of models, tasks, and objectives receive this search budget. Final selected rates, ranks and checkpoints may differ between objectives.

Immediate training targets the edited frame; delayed training targets four subsequent autonomous frames. Crafter and Cartpole weight mean squared error in the target region by 2 and preservation outside the target by 0.1. CoinRun uses the whole image as the target region, with weight 2; its spatial target leaves no term outside the target. The WM remains frozen.

\textbf{Editor implementations and fitting budgets.}
The nonlinear MLP uses two hidden layers of width 64 with tanh activations. Its standardized input is represented in the complete numerical span of the training input rows, with no tuned dimensionality truncation. Training minimizes source-weighted MSE of coefficients in the original rank-eight correction basis. AdamW \citep{loshchilov2019decoupled} uses learning rate $10^{-3}$, weight decay $10^{-3}$, gradient clipping at 10, and 512 updates; validation coefficient MSE selects among updates 16, 64, 128, 256 and 512. The seed is 71901. DIAMOND fits one head per denoiser call.

Manifold stores paired training reference activations as anchors. A symmetric Euclidean nearest-neighbor graph \citep{cover1967nearest} starts at $k=6$ and increases $k$ only until connected. The destination minimizes standardized target-semantic squared discrepancy plus 0.1 times squared normalized recipient distance. After attaching the recipient to its nearest anchors, a shortest path \citep{dijkstra1959note} defines the correction. The traveled arc length is $\min(\alpha L,\rho)$ for total path length $L$. Fitting constructs the graph and normalization statistics without gradient updates. Sparse anchors and possible single-segment paths limit claims about smooth or curved manifold recovery.

For Concept DAS (CDAS), let $F(c)$ denote the frozen model's edited-frame output when the intervention activations are replaced by $c$ in the recipient context, and let $D_R$ denote a norm-capped distributed interchange intervention (DII). Its paired-teacher loss is
\begin{equation}
\mathcal L_{\rm CDAS}=\tfrac12\bigl[\operatorname{JSD}(Q(F(D_R(a,a^\star))),Q(F(a^\star)))
+\operatorname{JSD}(Q(F(D_R(a^\star,a))),Q(F(a)))\bigr].
\end{equation}
Here JSD denotes the Jensen--Shannon divergence \citep{lin1991divergence}. The fixed readout $Q$ applies a normalized Gaussian soft quantization independently to each decoded red--green--blue (RGB) channel using 16 bins on $[0,1]$ and standard deviation $1/15$. It is a proxy distribution, not the full predictive image distribution used as an ideal distribution-matching target. Teachers are detached; only the edited frame enters this loss. Rank-eight subspaces initialize from training correction PCA and are orthonormalized by QR decomposition \citep{householder1958unitary}. Adam \citep{kingma2015adam} uses learning rate $10^{-3}$, gradient clipping at 1, 256 updates, and seed 71903; sources and then their events are sampled uniformly. Validation bidirectional JSD selects among updates 32, 128 and 256; update 0 is diagnostic only. Gradients use native-forward straight-through implementations through frozen model operations. A source-weighted ridge map with penalty $10^{-3}$ then predicts $Ra^\star$ from standardized $[a;r]$ using training pairs. Inference applies $R^\top(\widehat c-Ra)$ under the selected dose and original norm cap. This conditional clamping and soft-pixel readout adapt the original Concept DAS objective to the repair setting; they do not demonstrate isolation of a named concept.

For the nonlinear MLP, Manifold, and Concept DAS, the final dose maximizes source-equal validation SSG@8 over $\{0.03,0.1,0.3,1\}$. Training events number 15, 36 and 23 for Crafter, Cartpole and CoinRun; validation events number 6, 12 and 25. The corresponding training/validation source counts are 15/5, 18/6 and 8/8. CoinRun uses disjoint development levels, so validation tests cross-level transfer. All methods retain the source-disjoint splits and intervention sites, but their training compute and hyperparameter searches are not matched. Each new gradient-trained method uses one training seed. The graph anchors and paired supervision make these small-data baselines; their results cannot establish the headroom of fully data-scaled editors. New confidence intervals use 10,000 source-bootstrap draws \citep{efron1979bootstrap}, recomputed from the audited event reports to match the paper's resampling budget, and remain conditional on the fitted editors.

\subsection{World-model training and checkpoints}\label{app:wm-training}
Frames are $64\times64$ RGB. The shared history allowance is 64 frames; DIAMOND, DreamerV3 and STORM consume four, 64 and eight, respectively. Native sampling and intervention parity audits bind the frozen checkpoints and evaluation code. STORM Cartpole uses continuous requested actions without discretization. The offline runs for DIAMOND and STORM on Cartpole use the same replay pool of 110,004 frames.

The training budgets in Table~\ref{tab:training-budgets} follow each WM's native accounting. For online DreamerV3 training, one \emph{decision} is one environment interaction step in which the agent selects an action and receives the next transition; model and policy optimization occurs during this interaction process according to the native training ratio. For offline DIAMOND and STORM training, one \emph{update} is one optimizer step on a minibatch sampled from the fixed replay. Decisions and updates therefore document how each checkpoint was obtained but are not directly comparable units of training compute.

\input{tables/training_budgets.tex}

STORM Cartpole resumed from a verified model and Adam state at 10,000 updates and reached 20,000 updates. Sampling of replay indices was reconstructed; Torch sampling restarted with seed 8102 and the scaler for mixed precision was reset. The final checkpoint was fixed before confirmation.

CoinRun uses the easy distribution. DreamerV3 collected 30000 training frames across 179 distinct levels sampled from indices 0 to 199. DIAMOND and STORM train on these same images and actions, with 28106 eligible short windows and 9500 windows of 64 steps without resets, respectively. Editor fitting uses levels 10000 to 10007 and validation uses levels 10008 to 10015. All three WMs share the resulting 48 development events and the independent confirmation levels. The terrain readout and editor selection use only development data.

\subsection{Additional diagnostic protocol}\label{app:diagnostic-protocol}
The RAP control study rescales the RAP direction to each included fitted editor's realized norm without fitting or selection on confirmation data. The restoration study across all models reuses the first frozen confirmation event from every source, all three seeds and fixed schedules of actions and noise over 24 steps. It evaluates RAP, Immediate LoReFT, and the separately reported Delayed LoReFT unchanged and after restoring the untouched latent, memory, or both at step 1. For DIAMOND, which has no recurrent pair of hidden state and cache, the registered architectural analogue defines the newest generated frame as the latent component and the preceding three conditioning frames as memory. Joint restoration exactly recovers the paired untouched future in every audited branch.

\section{Complete Intervention Results and Uncertainty}\label{app:statistics}
Statistical confidence intervals (CIs) use 10,000 bootstrap draws over sources. Sampling seeds and events are averaged within a source before resampling independent sources. The common reference mask is shared across methods. Tables~\ref{tab:confirmation-diamond} to~\ref{tab:confirmation-storm} combine ISG, SSG@8 and endpoint results for each model, avoiding a second roster of the same interventions. Endpoint gain measures the paired semantic effect at $h=8$, whereas SSG@8 averages scoreable autonomous frames within each event before source aggregation.

\subsection{Model-wise intervention profiles}\label{app:modelwise-results}
The three complete tables are organized by architecture, with a focused interpretation immediately introducing each numerical block. Read across a row to compare immediate repair, average retained benefit, and the final measured step; read within a task block to compare fitted editors under the same sources, events, actions, and noise. This layout makes temporal disagreement visible without duplicating the estimates in separate metric-specific tables.

DIAMOND displays both substantial interface headroom and strong variation across time. RAP yields SSG@8 gains of 10.76, 9.01 and 4.14 pp on Crafter, Cartpole and CoinRun. At the endpoint these become 8.02, 12.71 and 1.47 pp: Crafter and CoinRun lose part of the average benefit by the last step, whereas Cartpole strengthens. The fitted rows also separate immediate from persistent behavior. Delayed LoReFT reaches 3.24 pp SSG@8 on Cartpole despite only 0.34 pp ISG, but becomes harmful on CoinRun; the normalized probe on CoinRun has a positive sustained interval while its endpoint interval includes zero. Thus neither the edited frame nor the endpoint alone recovers the ordering induced by the full autonomous window.

\input{tables/combined_statistics_diamond.tex}

DreamerV3 Cartpole is the only combination where fitted editors exceed RAP: Immediate and Delayed LoReFT reach 3.22 and 3.46 pp SSG@8, compared with 0.92 pp for RAP. Their endpoint gains rise further to 3.91 and 4.07 pp, so this reversal is not produced by a transient early advantage. On Crafter and CoinRun, RAP remains larger than every fitted method; moreover, most fitted Crafter intervals include zero, while CoinRun improvements remain small. The recurrent interface can therefore support a transferable learned direction, but only in one task rather than as a model-wide property.

\input{tables/combined_statistics_dreamerv3.tex}

For STORM, the inverse probe on Crafter and normalized probe on CoinRun retain positive sustained intervals. Immediate LoReFT is harmful on Crafter, whereas the nonlinear MLP is positive on Cartpole, demonstrating that a method family does not inherit one direction of effect across tasks. The other effects are smaller or uncertain, while RAP remains positive on all three tasks. Across the three architecture tables, RAP has positive ISG and SSG@8 intervals in all nine combinations, so accessible corrective information is much more consistent than fitted recovery of it.

\input{tables/combined_statistics_storm.tex}
\clearpage

\subsection{Scoreability and endpoint retention}\label{app:coverage-retention}
The endpoint comparison above is meaningful only if later horizons do not selectively discard difficult events. Table~\ref{tab:scoreable-event-coverage} therefore reports the common event denominator before any model or method comparison. Coverage is effectively complete: Cartpole and CoinRun retain every event at every horizon; Crafter retains all 41 events except for one exclusion at $h=1$ and one at $h=5$, and returns to the full denominator at $h=8$. The endpoint mean for RAP is below SSG@8 in eight of nine model--task combinations, while its endpoint confidence interval remains above zero in eight; DIAMOND CoinRun is the exception. Because the final horizon does not suffer lower coverage, this widespread attenuation reflects changes along the generated trajectories rather than systematic removal of difficult late frames.

\input{tables/scoreable_event_coverage.tex}

\subsection{Interface headroom and the DreamerV3 exception}\label{app:special-cases}
\input{figures/special_cases.tex}
Figure~\ref{fig:special-cases}a visualizes the interface claim based on the table in Section~\ref{sec:results-patch}; it is placed here because the main text does not need a second presentation of the same nine values. Panel b isolates the sole exception without presenting it as the typical result. Immediate LoReFT reaches 3.22 pp, compared with 0.92 for full RAP, 1.18 for RAP matched by norm, and 0.24 for its random edit matched by norm. The paired difference between RAP and Immediate LoReFT is $-2.30$ pp with 95\% CI [$-2.68,-1.90$] under a bootstrap over sources. Table~\ref{tab:dreamer-exception-seeds} further shows the same ordering under every registered prediction seed. These seeds reuse the same 48 sources and are robustness strata, not three new independent datasets. The evidence therefore supports a stable advantage of the learned direction in this cohort, while not identifying which recurrent feature creates it.
\input{tables/special_case_robustness_v5.tex}

The seed-wise table tests whether the exceptional cohort depends on one prediction-noise draw; the broader paired analysis below instead asks whether the observed winner in each cohort remains separated after source resampling and simultaneous adjustment.
\input{tables/paired_ranking_ci.tex}

After adjustment across the nine cohort comparisons, seven intervals retain a RAP advantage over the strongest fitted editor. DreamerV3 on Cartpole favors the fitted editor, while the RAP comparison for STORM on Crafter includes zero. This isolates one robust reversal and one unresolved ordering rather than treating small point estimate differences as settled rankings.

\section{Transfer and Random Controls}\label{app:transfer-controls}
\subsection{Matched random directions and requests with zero change}\label{app:matched-random-controls}
Each fitted editor has a random intervention matched to its realized norm at every edit call. Table~\ref{tab:intervention-diagnostics} summarizes the two highest and lowest SSG@8 fitted editors, using a fixed order for ties. Requests with zero change ask fitted editors to preserve the descriptor of the predicted image. They test unsolicited edits under the same generated state; the scorer still compares predictions with real observations. Thus a negative effect under a request with zero change means reduced quality against the real reference, and a nonzero edit need not coincide with a change in semantic score.

DreamerV3 controls are calibrated on the native bfloat16 grid. Scalar adjustment follows the prescribed random or reference direction; when quantization skips the requested norm, individual coordinates move by one representable step while retaining direction signs. Calibration uses activation geometry alone. Every realized control norm must match its target within 0.5\% plus $5\times10^{-4}$. Paired execution verifies exact replay without an edit, unchanged prediction prefixes and preserved recipient state.
\input{tables/intervention_diagnostics.tex}

\subsection{Complete validation, test and random control matrix}\label{app:complete-control-matrix}
Random controls exist only on the test events. Tables~\ref{tab:all-random-controls-panel-diamond} to~\ref{tab:all-random-controls-panel-storm} report all 99 fitted editor--cohort rows, including Delayed LoReFT for supplementary completeness. Validation and test columns measure transfer to new sources, while the final contrast compares each fitted direction with a random direction of the same realized magnitude. Each model table places the three tasks side by side and separates immediate from sustained effects.

\input{tables/random_controls_panel_diamond.tex}

For DIAMOND, the normalized probe on CoinRun and Delayed LoReFT on Cartpole retain the clearest sustained test effects above random. Delayed LoReFT on CoinRun instead changes from positive validation SSG@8 to a harmful test effect and falls 1.16 pp below its matched random control.

\input{tables/random_controls_panel_dreamerv3.tex}

DreamerV3 Cartpole provides the strongest learned transfer. Immediate and Delayed LoReFT exceed matched random controls by 2.98 and 3.36 pp SSG@8. Most Crafter effects remain small, and the positive CoinRun effects are substantially weaker than the Cartpole results.

\input{tables/random_controls_panel_storm.tex}

STORM separates probe transfer from LoReFT failure. The inverse probe on Crafter and normalized probe on CoinRun exceed random by 2.27 and 0.62 pp SSG@8. Immediate LoReFT on Crafter falls 1.38 pp below random, and Delayed LoReFT is harmful on Crafter and CoinRun. For the nonlinear MLP, Manifold, and Concept DAS, all three DreamerV3 Cartpole directions exceed random by 0.39--0.70 pp SSG@8 with pointwise source-bootstrap intervals above zero; DreamerV3 CoinRun also favors Manifold and Concept DAS by 0.03 and 0.28 pp. Nonlinear MLP exceeds random by 0.59 pp on STORM Cartpole, whereas Manifold falls 2.95 pp below random on DIAMOND Cartpole. Across the 90 rows in the expanded fitted comparison, 68 have lower test SSG@8 than validation and 25 reverse sign. Across all fitted editors, equal magnitude therefore neither explains successful transfer nor prevents a harmful direction.

\section{Magnitude and Direction Controls}\label{app:magnitude-direction}
\subsection{Reference direction at fitted magnitude}\label{app:reference-at-fitted-magnitude}
The available RAP controls matched by norm rescale the RAP correction to the realized norm of each fitted editor included in the control study at every edit call. DIAMOND matches its three denoiser calls separately. Events, actions and prediction noise are shared. Table~\ref{tab:matched-patch} gives the LoReFT comparison; the audited values for the fitted editors in panels a--c of Figure~\ref{fig:editor-gaps} accompany the figure data. The nonlinear MLP, Manifold, and Concept DAS have no such controls.
\input{tables/matched_patch_comparison.tex}

\subsection{Fitted direction at reference magnitude}\label{app:fitted-at-reference-magnitude}
The reverse control keeps the Delayed LoReFT direction and rescales it to the realized RAP norm. Together with RAP at the fitted norm, this tests both directions under the other method's magnitude. Increasing the fitted direction to the reference magnitude does not consistently recover the RAP effect and can amplify harm.
\input{tables/matched_delayed_full_norm.tex}

\subsection{Reference timing}\label{app:reference-timing}
For DreamerV3 Crafter, shifting reference images eight steps earlier changes RAP ISG from 2.71 to $-0.48$ and SSG@8 from 1.24 to 0.42. Endpoint gains fall from 0.74 to 0.40. An outdated reference can therefore harm the edited frame while retaining a smaller later benefit. Edit norms averaged by source are 73.64 and 73.47; individual event norms remain unconstrained. This control measures sensitivity to reference age under full replacement.

\section{Temporal Robustness and Long Horizon Behavior}\label{app:temporal}
\subsection{Sensitivity to horizon, action and prediction noise}\label{app:temporal-sensitivity}
The following analyses reuse frozen confirmation reports and never count action strata or prediction seeds as new independent datasets. They compare all eleven fitted editors, including Delayed LoReFT, and are supplementary to the baseline table. For $H\in\{1,2,4,8\}$, an event requires at least $\lceil0.75H\rceil$ scoreable frames after editing. Action sensitivity stratifies the registered future action window of eight steps without changing it: Cartpole and Crafter use lower versus higher action variation at the task median, while CoinRun uses windows with constant versus switched held actions. Noise sensitivity recomputes every cohort separately for each registered seed.
\input{tables/horizon_sensitivity.tex}

Method rankings depend on when they are measured. The fitted leader at $H=8$ differs from the leader at $H=1$ in four of nine combinations. RAP also has no universal horizon profile. Its gain falls from 12.81 to 10.76 pp for DIAMOND on Crafter, but rises from 6.00 to 9.01 pp for DIAMOND on Cartpole. A single horizon therefore cannot summarize either fitted editor quality or the correction available at an interface.
\input{tables/action_sensitivity.tex}

The future action sequence changes the fitted leader in five of nine combinations. Reference headroom remains positive in every action stratum, but its size can change substantially. For STORM on Crafter, RAP rises from 2.24 pp under lower action variation to 4.81 pp under higher variation. The value carried by the same interface therefore depends on the trajectory that follows the edit.
\input{tables/noise_sensitivity.tex}

Prediction noise changes the fitted leader in four of nine combinations. RAP remains positive for every registered seed, which makes the presence of reference headroom more stable than the identity of the best fitted editor. The variation across seeds is retained in the table because averaging alone would hide these ranking changes.

\subsection{Representative long-horizon regimes}\label{app:long-horizon-regimes}
Figure~\ref{fig:temporal-regimes} extends the comparison to 24 prediction steps. A benefit can decay while remaining positive, emerge only after the edited prediction, or remain harmful throughout later prediction. These regimes explain why immediate repair, average retained gain and endpoint behavior must be reported separately.

\input{figures/temporal_regimes_figure.tex}\clearpage

\section{Metric Validation and Qualitative Examples}\label{app:diagnostics}\label{app:cases}
\subsection{Prediction and measurement competence}\label{app:measurement-competence}
A completed rollout and an executable edit establish implementation feasibility. Task qualification additionally checks whether the generated target is recognizable and whether the metric responds to its errors. These checks precede interpretation of tool rankings.

Crafter scores are based on calibrated tile measurements. Calibration on real images and recognition of objects in generated images remain different validation questions. Aggregate gains weight tile classes and should be interpreted with their class coverage. Cartpole scores measure visible overlap of body masks. These observables specific to each task support matched tool comparisons within each task; their numerical scales do not imply equal difficulty across tasks or architectures.

CoinRun measures the geometry of empty space and solid platforms. Native simulator state supplies reference labels for scoring; projected tile borders, entity occlusions and unsupported terrain are masked. A frozen convolutional readout labels generated and real images. Pixels enter the common mask when the label from the real image matches the reference; each scored frame requires at least 90\% reference coverage and 32 retained pixels of each class. Mean class recall on eight held out development levels is 0.999846. Manual review of the generated frames and controls based on spatial shifts checks the readout on model outputs. Across the 16 development levels, baseline terrain accuracy is 0.821 for DIAMOND, 0.942 for DreamerV3 and 0.907 for STORM. DIAMOND exhibits camera drift and has lower average quality than holding the last real observation fixed. These are qualification diagnostics; the formal tool comparison uses independent levels. At evaluation, simulator labels and future real frames are accessible only to the scorer. Editors receive targets derived from the current image with descriptors of 128 dimensions.

\subsection{Task overview examples}\label{app:task-examples}
Figure~\ref{fig:task-composition} uses the first event from each of the first two distinct sources in the frozen task rosters. Cartpole examples additionally require at least 30 pixels of the reference body and a bounding box within pixels 2 to 61, so the body is visible. This illustration rule leaves every test source and event in the reported aggregates. Images are unchanged real observations at the editing time.

\subsection{Fixed native example}\label{app:fixed-native-example}
Figure~\ref{fig:teaser} uses STORM level 30011, prediction root 105, an edit after four predictions and the first registered noise seed, 69000. The event was selected for a visible current repair followed by declining accuracy; aggregate estimates retain the full cohort. RAP gives ISG 24.25 and SSG@8 14.73 on this event. Its positive endpoint gain of 8.72 coexists with a drop in absolute terrain accuracy from 96.44\% to 69.98\%. The image strip enlarges the original $64\times64$ RGB frames at the edited time and after eight steps, with adjacent panels separated by black borders. Both curves retain all nine times: absolute terrain accuracy on the left and the gain of edited over untouched predictions on the right.

\section{Component Restoration and Pathway Analysis}\label{app:temporal-pathways}
\subsection{Protocol and complete aggregate results}\label{app:pathway-protocol}
The extended restoration study covers all three models and all three tasks for RAP, Delayed LoReFT and Immediate LoReFT. It retains the first frozen event from each source and all three prediction seeds, giving 1,197 combinations of models, events, and seeds before method and restoration branching. Models, editors, actions and sampling schedules remain fixed. Each window of eight steps requires at least six scoreable reference frames. This leaves 37 Crafter, 48 Cartpole and 47 CoinRun sources per model for the common pathway analysis over steps 9 to 24.

At step 1, we restore selected components from the paired untouched prediction. For DreamerV3 and STORM, write the native state as $(z_1,m_1)$, with $z_1$ the sampled latent and $m_1$ the deterministic recurrent state or complete transformer cache. For DIAMOND, $(z_1,m_1)$ denotes the newest generated context frame and the preceding three context frames. Superscripts $\mathrm{base}$ and $\mathrm{edit}$ identify untouched and edited branches. Every method evaluates
\begin{equation}
 (z_1^{\mathrm{base}},m_1^{\mathrm{edit}}),\qquad
 (z_1^{\mathrm{edit}},m_1^{\mathrm{base}}),\qquad
 (z_1^{\mathrm{base}},m_1^{\mathrm{base}}).
\end{equation}
All branches preserve the original edited frame at step 0. Audits of saved frames verify the replaced and retained components. Joint restoration exactly recovers every untouched future frame. Means across all models appear in Table~\ref{tab:v4-restoration-all}; the figure in the main text shows selected mean trajectories without confidence shading.

\begingroup\scriptsize
\input{tables/restoration_all.tex}
\endgroup

\subsection{Time resolved component pathways}\label{app:pathway-time-resolved}
Figures~\ref{fig:pathway-diamond} through~\ref{fig:pathway-storm} report restoration loss at every prediction step for all three tasks and all three intervention methods. The DIAMOND plots separate the newest generated frame from the earlier context frames. The DreamerV3 plots separate the sampled latent from recurrent memory, and the STORM plots separate the sampled latent from the transformer cache. These trajectories show when a component begins to carry an effect and whether that dependence persists, changes sign or disappears.

\input{figures/pathway_diamond.tex}
\input{figures/pathway_dreamer.tex}
\input{figures/pathway_storm.tex}\clearpage

\subsection{Interpretation and controls with hybrid states}\label{app:pathway-hybrid-controls}
Restoration loss $L_c=G(I)-G(R_c\!\circ I)$ is signed. Positive loss means that replacing an edited component with its untouched value removes later benefit; negative loss means that restoration improves the trajectory and the edited component was associated with harm. Neither sign alone identifies a pure semantic variable, because restoration of a single component creates a hybrid state and may alter compatibility between components.

Table~\ref{tab:pathway-loss} summarizes the later gain over steps 9 to 24 and the loss caused by restoring each component at step 1. It places the time resolved curves on a common source bootstrap scale and makes the architectural differences explicit.
\input{tables/pathway_statistics.tex}

The dominant component differs across architectures. On DreamerV3 Cartpole, Delayed LoReFT retains 4.95 pp and restoring memory removes 4.68 pp, while restoring the sampled latent removes only 0.18 pp. For STORM Cartpole, RAP depends on both components, with losses of 3.10 pp for the sampled latent and 2.80 pp for memory. On STORM Crafter, Delayed LoReFT is harmful and either restoration improves the later trajectory, implicating both components in the harmful displacement.

The earlier restoration study with matched random controls is available for DreamerV3 Cartpole, STORM Cartpole and STORM Crafter. It subtracts the restoration loss after a random edit with equal norm from the loss after the semantic edit:
\begin{equation}
 L_c(I_{\rm semantic})-L_c(I_{\rm random}).
\end{equation}
\input{tables/random_restoration_controls.tex}

Five of the twelve semantic minus random intervals exclude zero. The strongest are DreamerV3 memory for Delayed LoReFT at 4.62 pp and both STORM Cartpole components for RAP at 3.55 and 2.84 pp. The other intervals include zero. These controls support selective component dependence, not a complete causal decomposition. Since matched random restoration is unavailable for the other combinations, the broader patterns remain operational descriptions.

\section{Local Analysis and Interpretation Limits}\label{app:local-propagation}
\subsection{Local propagation conditions}\label{app:local-conditions}
Fix an event, its action sequence and prediction noise. Let $x_h$ and $\bar x_h$ be the untouched and edited predictive states, with $\delta x_h=\bar x_h-x_h$. Here $F_h$ denotes the native transition with fixed noise where it is smooth, or its differentiable local relaxation; derivative claims do not cover hard sampling boundaries. Assume that $F_h(\cdot;u_h,\xi_h)$ is twice continuously differentiable on the line segment between these states, with $\|D^2F_h\|_{\mathrm{op}}\leq M_h$. Write $J_h=D_xF_h(x_h;u_h,\xi_h)$ and define $\Phi_{h:k}=J_{h-1}\cdots J_k$, with $\Phi_{h:h}=I$. Taylor's theorem gives
\begin{align}
 \delta x_{h+1}&=J_h\delta x_h+r_h,
 &\|r_h\|&\leq \tfrac12 M_h\|\delta x_h\|^2,\label{eq:local-transition}\\
 \delta x_h&=\Phi_{h:0}\delta x_0+
 \sum_{k=0}^{h-1}\Phi_{h:k+1}r_k.\label{eq:local-unroll}
\end{align}
For a twice continuously differentiable quality surrogate $\widetilde q_h$, let $g_h=\nabla\widetilde q_h(x_h)$ and $\|D^2\widetilde q_h\|_{\mathrm{op}}\leq K_h$. A second Taylor expansion yields
\begin{equation}
 \Delta\widetilde q(h)=g_h^\top\Phi_{h:0}\delta x_0
 +g_h^\top\sum_{k=0}^{h-1}\Phi_{h:k+1}r_k+s_h,\qquad
 |s_h|\leq\tfrac12K_h\|\delta x_h\|^2.\label{eq:quality-remainder}
\end{equation}
For fixed finite horizon and trajectories that remain in these neighborhoods, the last two terms are $O(\|\delta x_0\|^2)$. The leading term in Equation~\ref{eq:quality-remainder} is therefore an explicit local approximation, not an assumption of globally linear dynamics.

\subsection{Direction, magnitude and ranking reversal}\label{app:direction-ranking}
Set $\delta x_0=\rho d$ with $\|d\|=1$, and define $b_h=\Phi_{h:0}^\top g_h$. For $V=\sum_{h=1}^H v(h)$ and $b_{\rm S}=V^{-1}\sum_{h=1}^H v(h)b_h$, the surrogate gains satisfy
\begin{equation}
 \Delta\widetilde q(0)=\rho\langle g_0,d\rangle+O(\rho^2),
 \qquad \widetilde{\ssg}@H=\rho\langle b_{\rm S},d\rangle+O(\rho^2).\label{eq:isg-ssg-direction}
\end{equation}
Thus two directions with equal norms, $d_1,d_2$, can have different sustained effects. In particular,
\begin{equation}
 \langle g_0,d_1-d_2\rangle<0,\qquad
 \langle b_{\rm S},d_1-d_2\rangle>0\label{eq:ranking-reversal}
\end{equation}
is a sufficient local condition for method 1 to rank below method 2 by immediate gain but above it by sustained gain. This formalizes the ranking reversal observed for the two STORM Cartpole objectives. Moreover,
\begin{equation}
 |\Delta\widetilde q(h)|\leq
 \rho\|g_h\|\prod_{j=0}^{h-1}\|J_j\|_{\mathrm{op}}+O(\rho^2).\label{eq:propagation-bound}
\end{equation}
If the Jacobian norms are uniformly bounded by $\lambda<1$ and $\|g_h\|$ is bounded, the influence at first order decays geometrically; products with expanding or rotating directions can instead preserve, amplify or reverse it. Because $J_j$ is evaluated under $(u_j,\xi_j)$, action and noise changes alter this operator rather than merely adding measurement variance.

\subsection{State restoration and interpretation limits}\label{app:restoration-limits}
Let complementary projections $P_z$ and $P_m$ select the sampled latent and memory components at step 1. Restoring $z$ or $m$ changes the edited displacement to $P_m\delta x_1$ or $P_z\delta x_1$, respectively. For a later scoring window $W$, define $b_{W,1}=V_W^{-1}\sum_{h\in W}v(h)\Phi_{h:1}^\top g_h$. The corresponding losses in gain at first order are
\begin{equation}
 L_z=\langle b_{W,1},P_z\delta x_1\rangle+O(\|\delta x_1\|^2),
 \qquad
 L_m=\langle b_{W,1},P_m\delta x_1\rangle+O(\|\delta x_1\|^2).\label{eq:restoration-loss}
\end{equation}
Joint restoration sets $\delta x_1=0$ and recovers the untouched future under the same actions and noise. Interactions between $L_z$ and $L_m$ enter through higher order terms and can reflect off manifold state mixing. The matched random contrast estimates this generic mixing but does not make either component a pure semantic variable.

These derivations give local, testable conditions without assuming that a trained WM is contractive, globally linear or semantically factorized. Finite edit experiments establish operational pathway dependence; the equations state its interpretive limits.

\end{document}

%% file: figures/teaser.tex
\begin{figure}[!b]
\centering\includegraphics[width=\linewidth]{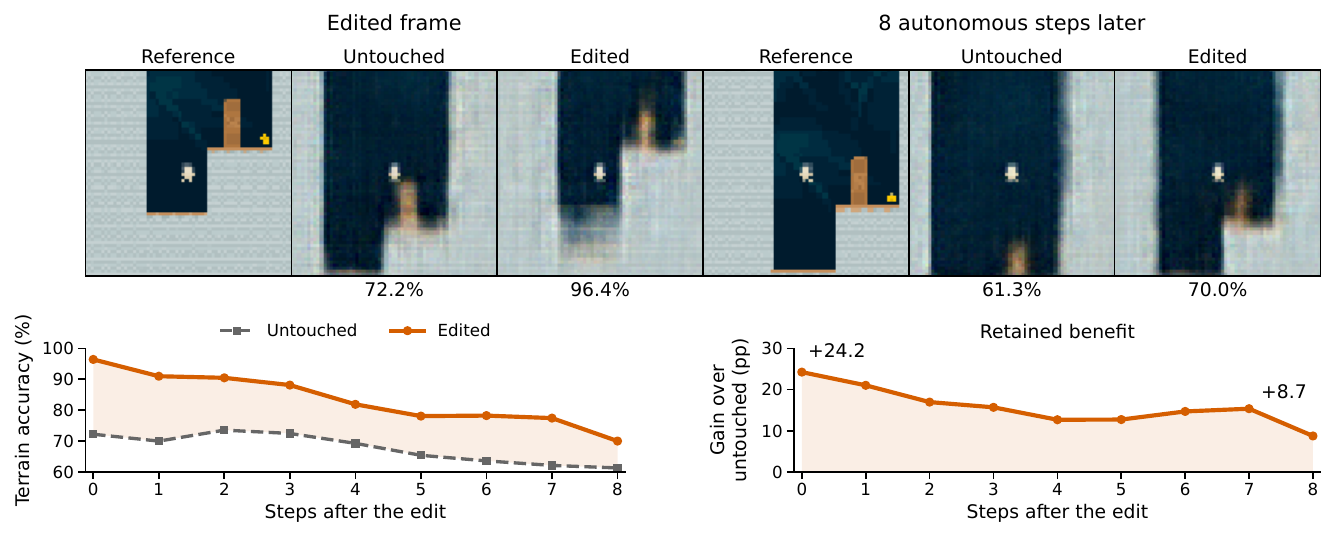}
\caption{\footnotesize\textbf{Current repair and later drift.} In STORM CoinRun, Reference Activation Patching (RAP) raises terrain accuracy from 72.2\% to 96.4\%. Eight steps later, accuracy is 70.0\% edited and 61.3\% untouched. Left: accuracy from 60 to 100\%; right: retained gain. Details in Appendix~\ref{app:cases}.}\label{fig:teaser}
\end{figure}

%% file: sections/01_introduction.tex
\section{Introduction}
World models (WMs) are used across environment simulation \citep{ha2018worldmodels,alonso2024diffusionworldmodelingvisual}, reinforcement learning and planning \citep{hafner2019learninglatentdynamicsplanning,schrittwieser2020muzero,hafner2024masteringdiversedomainsworld}, and synthetic data generation \citep{hu2023gaia1,jang2025dreamgen}. For a model that plans through imagined trajectories, the usefulness of a WM depends on how faithfully its predictions evolve over time. Dreamer uses recurrent latent dynamics \citep{hafner2024masteringdiversedomainsworld}, STORM combines stochastic latent states with a transformer \citep{zhang2023stormefficientstochastictransformer}, and DIAMOND generates visual trajectories through diffusion \citep{alonso2024diffusionworldmodelingvisual}. Despite their architectural differences, all three reuse generated states to predict the future, so both errors and corrections can evolve over many steps.

Prior work has discussed the difficulty of maintaining faithful trajectories \citep{hansen2026hallucinationworldmodelspredictable}. \citet{lu2026improvingweakworldmodels} document object and interaction failures in frozen Atari models, while \citet{vakalis2026interventiongaplatentworld} exposes differences between real and imagined intervention effects. Complementary benchmarks test whether predicted states remain consistent under changes to observations or commanded actions \citep{lu2026currentworldmodelslack,co2026worldsimprobediagnosingsimulatorfaithfulness}. Meanwhile, interpretability work probes which variables WMs encode \citep{zhang2026worldmodelslearnrl}, provides inspection interfaces indexed by time \citep{challagundla2026lensworldscapabilitytyped}, and demonstrates feedback steering \citep{hong2026steeringrobustnessworldaction}. These studies show that WMs can be inspected and edited, but they do not establish whether an internal correction remains useful after editing stops.

Answering this question requires separating current repair from future benefit. In Figure~\ref{fig:teaser}, Reference Activation Patching (RAP) raises terrain accuracy from 72.2\% to 96.4\% at the intervention step. After editing stops, accuracy falls to 70.0\% eight steps later but remains above the untouched prediction at 61.3\%. Evaluating only the current output would miss both the decay and the retained benefit.

We therefore propose \textbf{RolloutFaith}, a framework that separates two questions: Immediate Semantic Gain (ISG) asks whether the prediction at intervention time is repaired, whereas Sustained Semantic Gain (SSG) asks how much of that repair is retained over later autonomous predictions. We first use RAP as a diagnostic reference to test whether the selected interface can support a useful correction. Then we evaluate ten fitted editors adapted from probing and contrastive steering \citep{zhang2026worldmodelslearnrl,turner2024steeringlanguagemodelsactivation,zou2025representationengineeringtopdownapproach}, low rank latent carriers \citep{liu2026lowrankdynamicseffectivelatentcarriers}, and LoReFT \citep{wu2024reftrepresentationfinetuninglanguage}. To identify how an effect persists after editing stops, component restoration returns individual generated or memory components to their untouched values and measures how much of the gain disappears, revealing which states carry the correction forward. The evaluation spans DIAMOND, DreamerV3, and STORM on Crafter, DeepMind Control Suite (DMC) Cartpole, and Procgen CoinRun \citep{hafner2022benchmarkingspectrumagentcapabilities,tassa2018deepmindcontrolsuite,cobbe2020leveragingproceduralgenerationbenchmark}.

Across the nine model and task combinations, RAP improves SSG over eight future steps in all nine and exceeds the best fitted editor in eight, but its sustained gain decreases with horizon in five. This shows that the evaluated interfaces support useful corrections, while autonomous dynamics do not always preserve them. Fitted editors recover this capacity inconsistently because of limitations in correction magnitude, direction, and generalization, while component restoration reveals different persistence pathways across architectures. These findings expose a temporal objective mismatch: editors learn corrections at the intervention step, while success depends on their later consequences. We therefore propose \emph{Delayed LoReFT}, which optimizes through four frozen future transitions. It achieves its clearest sustained improvement on DIAMOND in Cartpole.

Our contributions are threefold:
\begin{enumerate}
\item \textbf{RolloutFaith framework and metrics.} We introduce ISG and SSG to measure both current repair and the benefit retained during autonomous prediction. Edited and untouched rollouts share events, actions, and prediction noise, enabling controlled comparison after intervention stops.
\item \textbf{Shared failure modes and component analysis.} We evaluate ten fitted editors across three WMs and three tasks and find that validation gains often fail to generalize to unseen test data. We show that the evaluated interfaces have substantial correction capacity, while fitted editors recover it inconsistently because of limitations in correction magnitude, direction, and generalization. Component restoration further reveals which state components carry intervention effects forward in each architecture.
\item \textbf{RolloutFaith improvement.} We introduce \emph{Delayed LoReFT}, extending LoReFT with temporal supervision through frozen future transitions. Its gains on several tasks provide direct evidence that optimizing future consequences can improve persistent intervention effects. This result opens a promising direction for developing interpretability tools that target the long term behavior of model predictions, rather than only the current output.
\end{enumerate}
\label{end:introduction}

%% file: sections/02_related_work.tex
\section{Related Work}\label{sec:related-work}
\subsection{Interpretability evaluation for world models}
Existing WM interpretability evaluations measure several different abilities. Atari probing tests whether internal states encode environment variables \citep{zhang2026worldmodelslearnrl}, and WorldModelLens provides interfaces for inspecting representations at different prediction steps \citep{challagundla2026lensworldscapabilitytyped}. WRBench measures whether outputs remain consistent when observations are hidden or revealed, while WorldSimProbe measures whether predicted states correctly realize commanded actions \citep{lu2026currentworldmodelslack,co2026worldsimprobediagnosingsimulatorfaithfulness}. The Intervention Gap tests whether the same environment intervention has consistent effects in real and imagined trajectories \citep{vakalis2026interventiongaplatentworld}. These evaluations therefore examine encoded information, observation consistency, action realization, and agreement between real and imagined effects. However, they evaluate the model at the queried step or compare outputs under external changes. They do not change an internal representation once and measure how that change affects future autonomous predictions.

\subsection{Intervention methods}
General methods relate internal representations to a model's current output. In visual diffusion, Revelio identifies semantic features across layers and denoising steps \citep{kim2025textitreveliointerpretingleveragingsemantic}, while Prompt Conditioned Intervention tests when inserted concepts persist through denoising \citep{gorgun2026temporalconceptdynamicsdiffusion}. These studies concern diffusion time within one generation, not future environment time. Activation Addition and Representation Engineering use contrastive directions \citep{turner2024steeringlanguagemodelsactivation,zou2025representationengineeringtopdownapproach}; AxBench benchmarks representation based steering \citep{wu2025axbenchsteeringllmssimple}; and Manifold Steering follows learned activation geometry \citep{wurgaft2026manifoldsteeringrevealsshared}. Activation patching and causal tracing restore activations \citep{meng2022locating,heimersheim2024useinterpretactivationpatching,zhang2024bestpracticesactivationpatching}, while DAS, LoReFT, and Concept DAS learn subspaces or objectives \citep{geiger2024findingalignmentsinterpretablecausal,wu2024reftrepresentationfinetuninglanguage,bao2026faithfulbidirectionalmodelsteering}. WM specific methods likewise target the current prediction: feedback steering applies repeated correction \citep{hong2026steeringrobustnessworldaction}, Concept-Guided Spatial Regularization (CGSReg) repairs task critical regions \citep{lu2026improvingweakworldmodels}, and low rank carriers infer corrections without references \citep{liu2026lowrankdynamicseffectivelatentcarriers}. None measures what remains after a single internal correction stops. RolloutFaith measures both current and future autonomous changes.

%% file: sections/03_framework_and_metrics.tex
\section{RolloutFaith: Evaluation Framework and Metrics}\label{sec:experiments}
RolloutFaith provides a unified paired-rollout framework for evaluating internal interventions at two timescales: their immediate correction of the edited prediction and the benefit retained during subsequent autonomous generation. By holding events, actions, and prediction noise fixed across edited and comparator branches, the framework supports controlled comparisons across models, tasks, and editing methods.

\subsection{Models and datasets}
\textbf{Models.} Three world models span complementary predictive mechanisms: DIAMOND generates images through diffusion \citep{alonso2024diffusionworldmodelingvisual}, DreamerV3 uses recurrent stochastic dynamics \citep{hafner2024masteringdiversedomainsworld}, and STORM represents latent histories with a transformer \citep{zhang2023stormefficientstochastictransformer}. Each exposes a native intervention state and supports reproducible paired sampling. Checkpoints and fitted editors are task specific, while evaluation events are shared across models within each task. This design tests persistence across distinct mechanisms for representing and propagating predictive state.

\textbf{Tasks.} Crafter, DMC Cartpole, and Procgen CoinRun satisfy two inclusion criteria: a usable intervention boundary and a calibrated image-space observable. Crafter measures tile semantics \citep{hafner2022benchmarkingspectrumagentcapabilities}, DMC Cartpole measures articulated geometry under continuous actions \citep{tassa2018deepmindcontrolsuite}, and CoinRun measures terrain on independently generated levels \citep{cobbe2020leveragingproceduralgenerationbenchmark}. For CoinRun, both criteria are verified for all three models on 16 development levels before test evaluation. Crossing tasks and models yields a shared $3\times3$ matrix spanning discrete and continuous actions, local and global observables, and generalization across episodes and procedurally generated levels.

\textbf{Dataset composition.} The test set contains 133 independent sources and 229 scoreable events: 38 Crafter episodes, 48 Cartpole episodes, and 47 CoinRun levels (Figure~\ref{fig:task-composition}). Common scoreability rules based on real references determine admission before replication across models, methods, or prediction seeds. All methods therefore share the same task-specific roster, while fitting, validation, and test data remain disjoint (Table~\ref{tab:task-coverage}). Results are first averaged over prediction seeds and events within each source, and then averaged equally across sources. This source-level aggregation prevents sources with more derived evaluations from receiving disproportionate weight.
\input{figures/task_composition_figure.tex}
\subsection{Intervention methods}\label{sec:tools}
To distinguish interface capacity from deployable editor performance, we first evaluate an idealized reference intervention. RAP replaces recipient activation $a$ with the event-matched reference $a^\star$: $a^{\prime}_{\mathrm{RAP}}=a^\star$. Because RAP accesses the test-event reference, it measures correction capacity at the selected interface rather than inference quality. RAP is therefore excluded from fitted-editor rankings and retains its native correction magnitude.

Fitted editors instead infer corrections without test reference activations, using training data for fitting and validation data for selection. To cover diverse correction mechanisms, we consider analytic direction, supervised parametric, and nonparametric geometric editors. Given request features $r$ and semantic residual $e=y^\star-g(a)$, method $m$ proposes $d_m(a,r,e)$ and applies
\begin{equation}
a^{\prime}_m=a+\operatorname{cap}^{(m)}_{\rho}\left(\alpha d_m(a,r,e)\right).
\end{equation}
Here $\rho$ is the 90th percentile of training correction lengths and $\alpha$ is selected on validation data. The method-aware cap limits Euclidean norm for additive proposals and arc length for path proposals, providing a common budget across correction geometries.

\textbf{Analytic direction editors.} Mean delta averages paired training corrections, while difference of means contrasts activations above and below the median semantic value. Inverse probe maps $e$ through the pseudoinverse of readout $P$; normalized probe combines unit readout directions. Because these probes use the task descriptors employed for evaluation, they assess targeted readout inversion. Caps apply at each native intervention call, including the three DIAMOND denoiser calls.

\textbf{Supervised parametric editors.} Principal component analysis (PCA) plus ridge, the uncentered carrier, and the nonlinear multilayer perceptron (MLP) predict coefficients in a fixed low-rank correction basis:
\begin{equation}
 d=\mu+Bf(z),\qquad z=\operatorname{standardize}([a;r]).
\end{equation}
PCA plus ridge and the carrier use $f(z)=Az+b$, fitted by ridge regression \citep{liu2026lowrankdynamicseffectivelatentcarriers}; centered PCA retains $\mu$, whereas the carrier sets $\mu=0$. The nonlinear MLP replaces this affine map with a two-layer tanh network trained on paired correction coefficients.

Immediate LoReFT \citep{wu2024reftrepresentationfinetuninglanguage} jointly learns an orthonormal subspace and request-conditioned affine map,
\begin{equation}
 d=R^\top(Wa+Ur+b-Ra),\qquad RR^\top=I,
\end{equation}
using image mean squared error (MSE) on the edited prediction. Concept DAS adapts bidirectional distribution matching \citep{bao2026faithfulbidirectionalmodelsteering} by learning an orthonormal subspace that matches frozen-model outputs in both exchange directions under Jensen--Shannon divergence (JSD). A training-only ridge map predicts target subspace coefficients from $[a;r]$. Appendix~\ref{app:implementation} gives full fitting details.

\textbf{Nonparametric geometric editor.} Manifold adapts Manifold Steering \citep{wurgaft2026manifoldsteeringrevealsshared} through a connected nearest-neighbor graph of training reference activations. Target semantics and recipient proximity select an anchor, and the correction follows the shortest graph path. For path length $L$, the cap traverses $\min\{\alpha L,\rho\}$ units. The resulting anchor graph represents empirical activation geometry through piecewise connections among observed training states.

\subsection{Evaluation metrics}\label{sec:metrics}
To measure immediate correction and post-intervention persistence, we propose Immediate Semantic Gain (ISG) and Sustained Semantic Gain (SSG). ISG evaluates the prediction produced under intervention ($h=0$); SSG averages improvement over subsequent autonomous predictions ($h\geq1$).

Task quality is body-mask intersection over union (IoU) for Cartpole, class-weighted tile accuracy for Crafter, and balanced terrain recall for CoinRun, each scaled to 0--100. Let $q_i^m(h)$ denote the quality of method $m$ on instance $i$ at horizon $h$, $C$ a comparator, and $v_i(h)$ the common scoreability indicator. For each comparator,
\begin{align}
\isg_C(m)&=\mathbb E_i[q_i^m(0)-q_i^C(0)],\\
\ssg_C@H(m)&=\mathbb E_i\left[\frac{\sum_{h=1}^H v_i(h)(q_i^m(h)-q_i^C(h))}{\sum_{h=1}^H v_i(h)}\right].
\end{align}
The expectation first averages seeds and events within each source and then weights sources equally. We use $H=8$, requiring a scoreable prediction at $h=0$ and at least six scoreable future frames. The scoreability mask is fixed across methods; Appendix~\ref{app:diagnostics} gives thresholds and coverage. Positive values indicate improvement over $C$, and negative values indicate harm. ISG measures immediate repair, whereas SSG measures benefit retained after editing stops. Because task metrics are not commensurate, values are compared only within a task.

\subsection{Evaluation protocol}
A common paired-rollout protocol instantiates the interventions and metrics. Each event fixes the source, checkpoint, edit time, request, interface, and actions; pairing an event with a prediction seed defines an evaluation instance. From a model-specific real-observation prefix, each model generates four unedited predictions, receives one intervention while producing $h=0$, and then runs autonomously for $h=1,\ldots,8$.

DIAMOND edits the final U Net up block at all three denoiser calls for the edited frame. DreamerV3 and STORM edit categorical prior logits while preserving recipient recurrent state or transformer history. Their real-observation prefixes contain four, 64, and eight frames, respectively.

Edited and comparator branches share events, actions, prediction noise, and scoreability masks under noise coupling \citep{ma2026twinrolloutsnoisecoupledcounterfactual}; only the intervention differs. Fitted editors use information available at edit time, while RAP alone receives the event-matched reference activation. Future real frames are reserved for scoring. Across the common roster, three models, and three seeds, the protocol yields 2,061 model--event--seed combinations. Appendix~\ref{app:implementation} specifies the interfaces and information budgets.

%% file: figures/task_composition_figure.tex
\begin{figure}[h]
\centering\includegraphics[width=.95\linewidth]{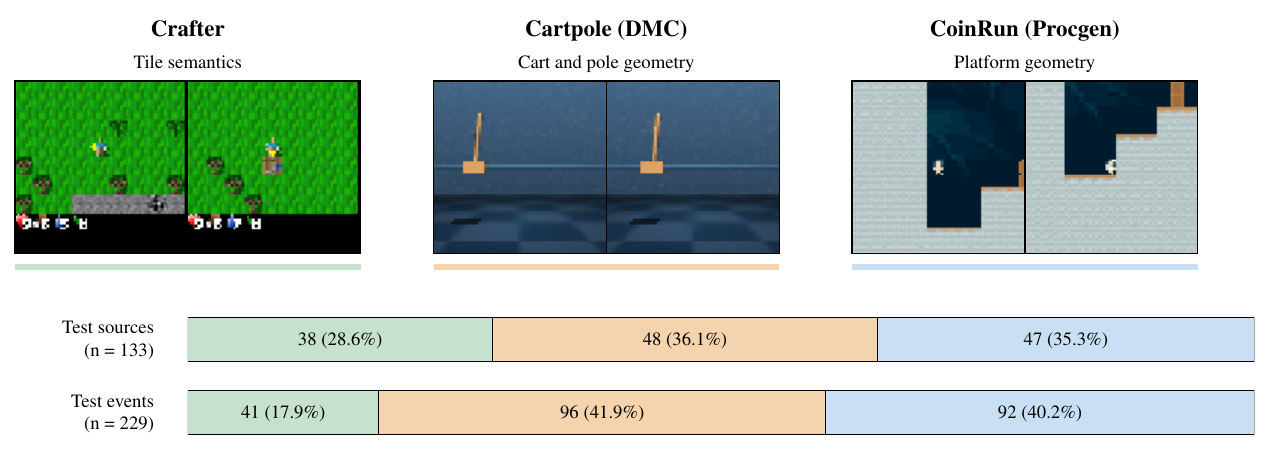}
\caption{\footnotesize\textbf{Task examples and test composition.} Real observations show each task. Sources are episodes for Crafter and Cartpole and levels for CoinRun; events are intervention locations.}\label{fig:task-composition}
\vspace{8pt}
\begin{minipage}{\linewidth}
\input{tables/task_coverage.tex}
\end{minipage}
\end{figure}

%% file: tables/task_coverage.tex
\centering\setlength{\tabcolsep}{2.3pt}
\makeatletter
\setlength{\belowcaptionskip}{5pt}
\def\@captype{table}
\caption{\footnotesize\textbf{Training, validation and test coverage}. Test pairs equal events times the three fixed seeds $\{69000,69001,69002\}$ per WM--method.}\label{tab:task-coverage}
\makeatother
\scriptsize
\begin{tabular}{@{}lrrrrrrr@{}}\toprule
& \multicolumn{2}{c}{Training} & \multicolumn{2}{c}{Validation} & \multicolumn{3}{c}{Test}\\
\cmidrule(lr){2-3}\cmidrule(lr){4-5}\cmidrule(l){6-8}
    Task & Sources & Events & Sources & Events & Sources & Events & Event--seed pairs\\\midrule
Crafter & 15 & 15 & 5 & 6 & 38 & 41 & 123\\
Cartpole (DMC) & 18 & 36 & 6 & 12 & 48 & 96 & 288\\
CoinRun (Procgen) & 8 & 23 & 8 & 25 & 47 & 92 & 276\\
\bottomrule\end{tabular}

%% file: sections/04_experimental_results.tex
\section{Experimental Results and Fitted-Editor Analysis}\label{sec:findings}
We conduct RolloutFaith experiments across three world models and three datasets to measure the ISG and SSG gains of RAP and fitted editors under the shared evaluation protocol. We analyze the reference--editor gap, the roles of correction direction and magnitude, and generalization ability.

\subsection{Reference--Editor Gap}\label{sec:results-patch}
\input{tables/main_matrix_v5.tex}

As shown in Table~\ref{tab:main-matrix}, RAP has positive SSG@8 in all nine model--task combinations, showing that every evaluated intervention interface supports a beneficial correction that persists after the intervention ends. RAP thus provides an informative idealized reference. However, its SSG@8 is lower than its ISG in most combinations, indicating that even reference-based corrections often yield smaller average benefits during subsequent autonomous prediction than at the intervention step.

A substantial gap remains between RAP and the fitted editors across nearly all model--task combinations, indicating that fitted editors recover the idealized correction capacity only partially and inconsistently. The sole exception is DreamerV3 Cartpole, where Immediate LoReFT outperforms RAP; Appendix~\ref{app:special-cases} analyzes this case. Most fitted editors nevertheless achieve positive gains in at least some settings. Many fitted-editor effects are small and their 95\% source bootstrap intervals include zero, so positive point estimates in Table~\ref{tab:main-matrix} do not necessarily indicate statistically resolved improvement. The primary limitation is therefore not the absence of useful corrections at the intervention interfaces, but the difficulty of estimating reliable correction rules from the available data. Future work should improve the data efficiency and cross-source generalization of fitted correction rules.

\subsection{Direction and Magnitude}\label{sec:results-direction}
An edit determines both a correction direction and magnitude, and either can limit its effectiveness. To separate their contributions, RAP matched by norm retains the RAP direction but scales it to the fitted tool's realized norm at each event and native edit call. This control preserves the reference direction under the fitted editor's correction budget. Comparing it with full RAP tests magnitude sensitivity, while comparing it with the fitted editor at the same magnitude tests direction quality. Further detailed results are provided in Appendix~\ref{app:magnitude-direction}.

\input{figures/editor_gaps.tex}
Figure~\ref{fig:editor-gaps}a isolates a failure driven predominantly by magnitude. For Immediate LoReFT on DIAMOND and Crafter, both the fitted edit and RAP matched by norm yield only small gains, whereas full RAP is substantially more effective. Replacing the fitted direction with the RAP direction at the same norm therefore recovers only a limited part of the gap; restoring the full RAP magnitude accounts for most of the improvement. This identifies insufficient scale as the dominant limitation in this case. Panel b instead isolates a direction failure for PCA plus ridge on DreamerV3 and Crafter. Full RAP and RAP matched by norm remain beneficial, while the fitted direction produces a negative gain. Because the matched control uses the fitted norm, its positive result shows that the available magnitude is sufficient; the reversal arises from the inferred direction. Together, panels a and b identify inadequate magnitude and inaccurate direction as distinct failure modes that aggregate gains alone cannot distinguish.

\subsection{Generalization}\label{sec:results-transfer}
To identify the source of negative test gains, Figure~\ref{fig:editor-gaps}c compares three representative fitted editors on validation and unseen test sources. Each comparison includes matched random edits and reference directions at the same norms. Positive validation gains become negative on unseen test sources, while the norm-matched reference directions remain beneficial and random edits have little effect. The decline is therefore not explained by the intervention interfaces or correction magnitudes alone; the fitted directions fail to transfer reliably to unseen sources.

To test whether these cases reflect a broader pattern, Figure~\ref{fig:editor-gaps}d extends the comparison to all 90 fitted combinations (Appendix~\ref{app:transfer-controls}). Test performance is below validation performance in 68 combinations, and 25 reverse from positive validation gain to harmful test gain. Although some combinations transfer successfully, the pattern recurs across fitted-editor types. It indicates a broad validation-to-test transfer gap under the available data budgets rather than an isolated failure of one editor type.

%% file: tables/main_matrix_v5.tex
\begin{table}[t]
\centering\normalsize
\setlength{\tabcolsep}{2.5pt}
\renewcommand{\arraystretch}{0.74}
\caption{\footnotesize\textbf{Intervention results under the shared evaluation protocol.} Point estimates are gains over untouched predictions in percentage points (pp), with equal source weights; all 95\% source bootstrap CIs are in Appendix Tables~\ref{tab:confirmation-diamond}--\ref{tab:confirmation-storm}. \textbf{Bold} marks the best fitted result per model, task, and metric; RAP is not ranked.}
\label{tab:main-matrix}
\begin{tabular*}{\linewidth}{@{\extracolsep{\fill}}llrrrrrr@{}}\toprule
& & \multicolumn{2}{c}{Crafter} & \multicolumn{2}{c}{Cartpole (DMC)} & \multicolumn{2}{c}{CoinRun (Procgen)}\\
\cmidrule(lr){3-4}\cmidrule(lr){5-6}\cmidrule(l){7-8}
WM & Tool & ISG & SSG@8 & ISG & SSG@8 & ISG & SSG@8\\\midrule
DIAMOND & Mean & +0.11 & \textbf{+0.31} & +0.09 & +0.06 & +0.06 & +0.14\\
 & InvProbe & \textbf{+0.89} & +0.21 & \textbf{+0.16} & -0.13 & +0.01 & +0.02\\
 & NormProbe & -0.23 & -1.06 & +0.01 & +0.02 & \textbf{+3.05} & \textbf{+2.25}\\
 & DiffMean & -0.43 & +0.09 & +0.00039 & +0.0040 & +0.34 & +0.16\\
 & PCA + ridge & -0.16 & -0.03 & -0.06 & +0.24 & +0.88 & +0.20\\
 & Carrier & -0.15 & -0.01 & -0.01 & +0.04 & +0.81 & +0.27\\
 & Immediate LoReFT & -0.50 & +0.18 & +0.02 & \textbf{+0.32} & +0.66 & +0.39\\
 & Nonlinear MLP & -0.07 & +0.08 & +0.04 & +0.07 & +0.04 & +0.09\\
 & Manifold & -0.20 & -0.11 & -0.67 & -2.94 & +0.48 & +0.34\\
 & Concept DAS & -0.08 & -0.07 & -0.18 & -0.14 & +0.40 & -0.23\\
\cmidrule(l){2-8}
 & RAP & +15.23 & +10.76 & +5.97 & +9.01 & +7.62 & +4.14\\
\midrule
DreamerV3 & Mean & +0.11 & -0.13 & +0.13 & +0.10 & +0.0044 & +0.0017\\
 & InvProbe & +0.28 & +0.19 & -0.16 & -0.03 & +0.03 & -0.07\\
 & NormProbe & -0.01 & \textbf{+0.32} & +0.19 & +0.54 & \textbf{+0.18} & +0.18\\
 & DiffMean & +0.06 & -0.04 & -0.35 & +0.03 & +0.0023 & +0.01\\
 & PCA + ridge & +0.11 & -0.15 & +0.88 & +1.02 & +0.10 & +0.37\\
 & Carrier & +0.12 & -0.10 & +0.74 & +1.09 & +0.09 & \textbf{+0.38}\\
 & Immediate LoReFT & -0.21 & -0.12 & \textbf{+2.57} & \textbf{+3.22} & +0.07 & +0.30\\
 & Nonlinear MLP & +0.05 & -0.02 & +0.70 & +0.76 & +0.00096 & +0.000055\\
 & Manifold & +0.03 & -0.04 & +0.62 & +0.60 & +0.02 & +0.01\\
 & Concept DAS & \textbf{+0.30} & +0.05 & +0.40 & +0.44 & +0.06 & +0.31\\
\cmidrule(l){2-8}
 & RAP & +2.71 & +1.24 & +0.52 & +0.92 & +1.06 & +1.20\\
\midrule
STORM & Mean & -0.87 & -0.03 & +0.22 & +0.37 & -0.01 & -0.04\\
 & InvProbe & \textbf{+2.17} & \textbf{+2.06} & -0.28 & +0.02 & +0.03 & -0.05\\
 & NormProbe & +0.62 & +1.44 & -0.03 & +0.21 & \textbf{+0.71} & \textbf{+0.44}\\
 & DiffMean & +0.24 & +0.09 & -0.13 & -0.07 & +0.05 & -0.02\\
 & PCA + ridge & -0.33 & +0.60 & +0.07 & +0.07 & -0.0038 & -0.01\\
 & Carrier & -0.17 & +0.17 & +0.0048 & +0.03 & -0.0040 & -0.01\\
 & Immediate LoReFT & -1.27 & -1.07 & +0.82 & +0.59 & +0.03 & +0.08\\
 & Nonlinear MLP & -0.66 & +0.22 & \textbf{+0.85} & \textbf{+0.66} & +0.01 & -0.05\\
 & Manifold & +0.13 & +0.24 & -2.78 & -1.16 & +0.06 & +0.04\\
 & Concept DAS & -0.04 & +0.09 & +0.08 & +0.05 & -0.01 & -0.03\\
\cmidrule(l){2-8}
 & RAP & +8.32 & +3.62 & +12.05 & +8.90 & +6.17 & +3.35\\
\bottomrule
\end{tabular*}
\end{table}

%% file: figures/editor_gaps.tex
\begin{figure}[t]
\centering\includegraphics[width=.98\linewidth]{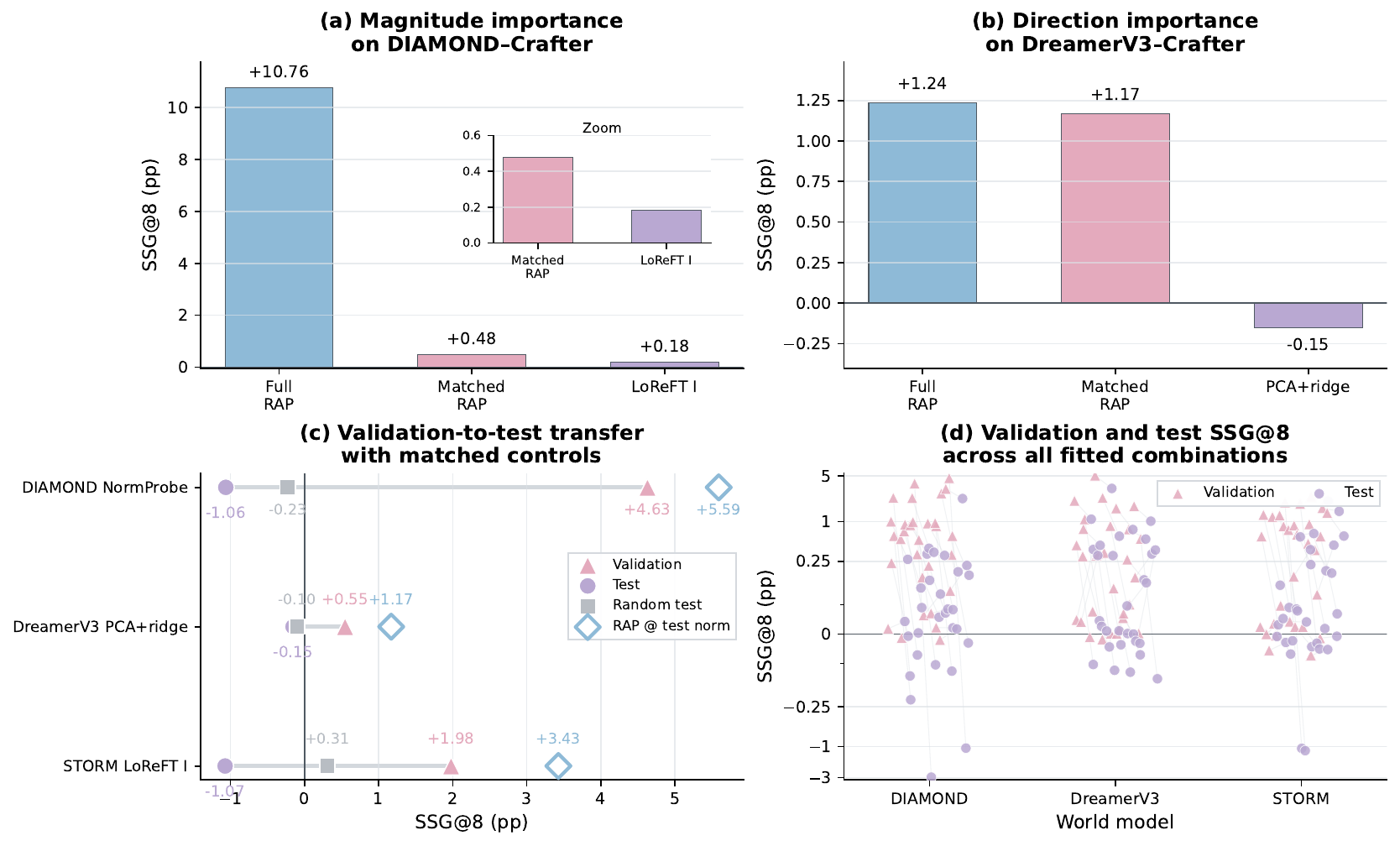}
\caption{\footnotesize\textbf{Direction, Magnitude, and Transfer Gaps of Fitted Editors} (a) Magnitude importance on DIAMOND--Crafter. (b) Direction importance on DreamerV3--Crafter. (c) Validation-to-test transfer with matched controls. (d) Validation and test SSG@8 across all fitted combinations.}\label{fig:editor-gaps}
\end{figure}

%% file: sections/05_component_analysis.tex
\section{Component Analysis}\label{sec:propagation}
To identify architecture-specific propagation pathways, we hold the original edit fixed, restore one candidate downstream component to its paired untouched value at step 1, and track the remaining effect under shared actions and noise. Differences across restorations indicate which components carry the edit into subsequent predictions. Appendix~\ref{app:temporal-pathways} details the 24-step restoration protocol, defines later gain $G$ as the average over steps 9--24, and reports the complete model--task matrix.

\input{figures/component_pathways.tex}
\paragraph{DIAMOND}
The restoration results in Figure~\ref{fig:component-pathways}a reveal a clear asymmetry: restoring the newest generated frame largely removes the later intervention effect, whereas restoring the preceding three-frame context has much less impact. DIAMOND conditions each prediction on a four-frame sliding visual context. Each new frame enters this context as the oldest is discarded, with no separate recurrent hidden state passed forward. The U Net edit affects the generated image, and its influence can pass into subsequent frames as the window advances. The newest frame carries the most recent manifestation of this influence, while the preceding frames provide earlier visual and temporal context. The restoration contrast indicates that the newest frame carries most of the persistent effect in this setting, although all four frames participate in prediction. This finding is consistent with DIAMOND's recursive use of generated images and supports visual context as its principal propagation pathway.

\paragraph{DreamerV3}
For DreamerV3, most of the later intervention effect survives restoration of the sampled latent but disappears when the deterministic recurrent state is restored (Figure~\ref{fig:component-pathways}b). Its recurrent state-space model (RSSM) combines deterministic state $h_t$ with stochastic categorical latent $z_t$. The recurrent transition updates $h_t$ from $h_{t-1}$, $z_{t-1}$, and the previous action; the updated state parameterizes the prior logits for $z_t$, and both components contribute to the current prediction. Our intervention changes these logits and the resulting sample while preserving the recipient $h_t$ at the intervention step. The edited latent affects the current frame and enters the next recurrent update, allowing its influence to accumulate in $h_{t+1}$. Restoring a later latent leaves this recurrent information intact, whereas restoring the deterministic state removes it. Thus, the stochastic latent introduces the correction into the dynamics, while the deterministic state becomes its principal long-term carrier. The restoration contrast supports this division of roles and is consistent with recurrent accumulation in the RSSM.

\paragraph{STORM}
Both restoration conditions substantially reduce the later intervention effect in STORM, although neither eliminates it; restoring either component also weakens harmful edits (Figure~\ref{fig:component-pathways}c). STORM predicts the next latent from the current sampled latent, action, and cached transformer history, updating the cache at each transition. The sampled latent is decoded into the current frame and passed to the next transition, providing a direct route for edited content to influence future predictions. The cache preserves information from earlier transitions, including the effects of previously edited latents, and provides a complementary historical route. Restoring either component leaves intervention-related information in the other, allowing some effect to persist. The observed reductions therefore implicate both components as substantial carriers in this setting, without establishing a single dominant pathway. This pattern is consistent with the joint use of current latent content and accumulated transformer history, supporting a distributed mechanism for intervention propagation.

%% file: figures/component_pathways.tex
\begin{figure}[!t]
\centering\includegraphics[width=.95\linewidth]{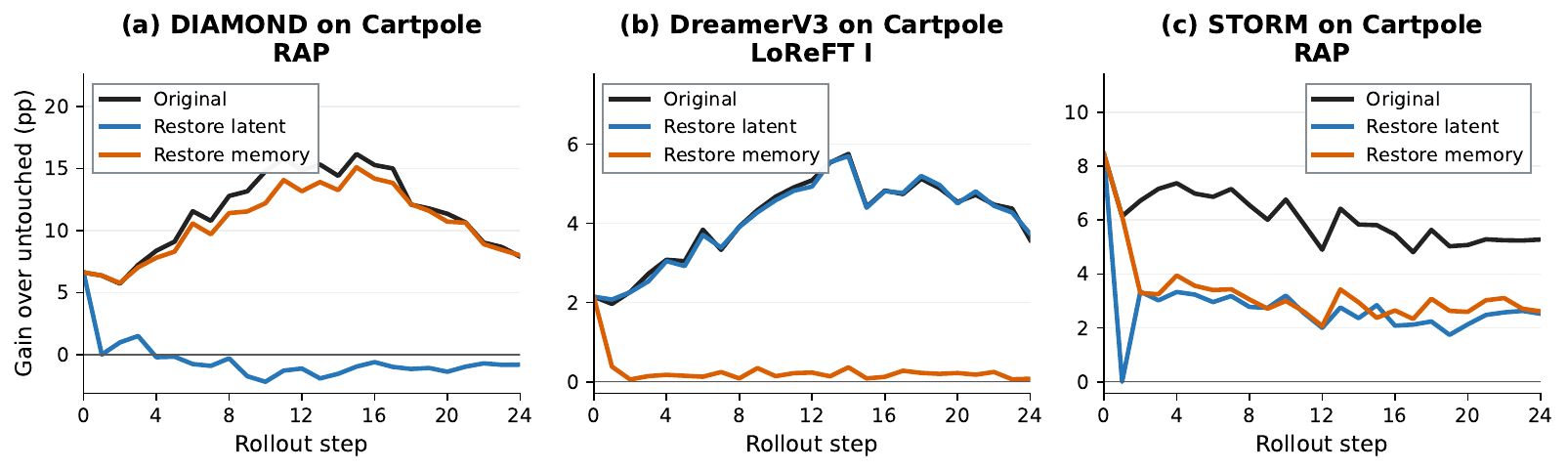}
\caption{\footnotesize\textbf{Intervention Propagation Pathways.} For a uniform legend, DIAMOND's ``latent'' and ``memory'' denote its newest generated frame and the context formed by the preceding three frames.}\label{fig:component-pathways}
\end{figure} 

%% file: sections/06_rolloutfaith_improvement.tex
\section{RolloutFaith Improvement}\label{sec:delayed-loreft}
\input{tables/delayed_loreft_v5.tex}

Rollout aware training extends generation beyond training horizons and reduces autoregressive exposure bias \citep{chen2024diffusionforcingnexttokenprediction,huang2025selfforcingbridgingtraintest}, motivating future transition supervision for an internal editor. We test this idea by unrolling frozen dynamics without changing LoReFT's evaluation interface.

Immediate LoReFT minimizes image loss on the edited prediction. Our \emph{Delayed LoReFT} instead unrolls the frozen model for $K=4$ registered transitions and minimizes future image loss:
\begin{equation}
\begin{aligned}
 \mathcal L_{\mathrm I}(\theta)&=\ell\!\left(\widehat o_t^{\,\mathcal I_\theta},o_t^\star\right),\\[-2pt]
 \mathcal L_{\mathrm D}(\theta)&=\frac{1}{K}\sum_{k=1}^{K}
 \ell\!\left(\widehat o_{t+k}^{\,\mathcal I_\theta},o_{t+k}^\star\right),\qquad K=4.
\end{aligned}
\end{equation}
Here $\mathcal I_\theta$ is the editor with parameters $\theta$, $\widehat o_{t+k}^{\,\mathcal I_\theta}$ and $o_{t+k}^\star$ are the predicted and real frames at step $t+k$, respectively, $\ell$ is image MSE, and $K$ is the training rollout horizon.

Table~\ref{tab:delayed-loreft} shows that \emph{Delayed LoReFT} outperforms Immediate LoReFT increasingly often as the evaluation horizon grows: $3/9$ cases for ISG, $5/9$ for SSG@8, and $6/9$ for final gain. This progression indicates that the advantage of rollout-based training becomes more apparent in later predictions, consistent with temporal supervision improving the persistence of an intervention rather than only its immediate effect. It therefore motivates supervising editors over longer prediction horizons so that training directly rewards the downstream consequences of an internal correction.

%% file: tables/delayed_loreft_v5.tex
\begin{table}[H]
\centering\small\setlength{\tabcolsep}{2.5pt}\renewcommand{\arraystretch}{0.82}
\caption{\footnotesize\textbf{Immediate versus Delayed LoReFT.} In the 24-step study, final gain $G_{17:24}$ is the mean paired improvement over untouched predictions from steps 17--24. Delayed results that improve on their Immediate counterparts are in \improvedcell{light blue} and marked with a green \improvementarrow.}
\label{tab:delayed-loreft}
\begin{tabular*}{\linewidth}{@{\extracolsep{\fill}}llrrrrrr@{}}\toprule
& & \multicolumn{3}{c}{Immediate} & \multicolumn{3}{c}{Delayed}\\
\cmidrule(lr){3-5}\cmidrule(l){6-8}
WM & Task & ISG & SSG@8 & Final gain & ISG & SSG@8 & Final gain\\\midrule
\multirow{3}{*}{DIAMOND} & Crafter & -0.50 & +0.18 & +0.31 & \improvedcell{-0.27\,\improvementarrow} & \improvedcell{+0.24\,\improvementarrow} & \improvedcell{+1.21\,\improvementarrow}\\
 & Cartpole & +0.02 & +0.32 & +0.01 & \improvedcell{+0.34\,\improvementarrow} & \improvedcell{+3.24\,\improvementarrow} & \improvedcell{+1.20\,\improvementarrow}\\
 & CoinRun & +0.66 & +0.39 & +2.29 & -8.67 & -1.17 & +1.13\\
\midrule
\multirow{3}{*}{DreamerV3} & Crafter & -0.21 & -0.12 & -0.16 & \improvedcell{-0.06\,\improvementarrow} & -0.17 & -0.23\\
 & Cartpole & +2.57 & +3.22 & +4.56 & +2.39 & \improvedcell{+3.46\,\improvementarrow} & \improvedcell{+4.89\,\improvementarrow}\\
 & CoinRun & +0.07 & +0.30 & +0.21 & +0.06 & +0.14 & \improvedcell{+0.67\,\improvementarrow}\\
\midrule
\multirow{3}{*}{STORM} & Crafter & -1.27 & -1.07 & -1.88 & -3.73 & \improvedcell{-0.60\,\improvementarrow} & \improvedcell{-1.63\,\improvementarrow}\\
 & Cartpole & +0.82 & +0.59 & +1.45 & -1.21 & \improvedcell{+0.75\,\improvementarrow} & +1.16\\
 & CoinRun & +0.03 & +0.08 & -0.26 & -1.51 & -1.06 & \improvedcell{+0.00\,\improvementarrow}\\
\bottomrule\end{tabular*}
\end{table}

%% file: sections/07_conclusion.tex
\section{Conclusion}\label{sec:conclusion}
We introduced \textbf{RolloutFaith} to evaluate the immediate and sustained effects of internal interventions under controlled paired rollouts. Reference Activation Patching reveals interface capacity; fitted editors are limited by correction magnitude, direction, and transfer, while component restoration identifies distinct propagation carriers across architectures. \emph{Delayed LoReFT} shows that future transition supervision can improve sustained effects. Future work should extend this supervision to longer horizons, develop correction inference without test references that transfers across unseen sources, and edit the state components that carry corrections so that their effects persist without repeated intervention.

%% file: tables/training_budgets.tex
\begin{table}[h]
\centering\small
\setlength{\tabcolsep}{3pt}
\caption{Recorded training budgets and batch layouts. Online training updates the WM and policy; offline training updates prediction modules on frozen Dreamer training replay. Batch layout gives sequences times frames per sequence.}\label{tab:training-budgets}
\begin{tabular}{lllrl}
\toprule
Task & WM & Training & Budget & Batch layout \\
\midrule
Cartpole & DIAMOND & Offline & 20,000 updates & $32\times5$ \\
Cartpole & DreamerV3 & Online & 110,004 decisions & $16\times64$ \\
Cartpole & STORM & Offline & 20,000 updates & $16\times64$ \\
Crafter & DIAMOND & Offline & 20,000 updates & $32\times5$ \\
Crafter & DreamerV3 & Online & 110,000 decisions & $16\times64$ \\
Crafter & STORM & Offline & 20,000 updates & $16\times64$ \\
CoinRun & DIAMOND & Offline & 5,000 updates & $16\times5$ \\
CoinRun & DreamerV3 & Online & 30,000 decisions & $16\times64$ \\
CoinRun & STORM & Offline & 5,000 updates & $16\times64$ \\
\bottomrule\end{tabular}
\end{table}

%% file: tables/combined_statistics_diamond.tex
\begin{table}[!htbp]
\centering\scriptsize\setlength{\tabcolsep}{2.2pt}
\caption{Complete intervention results for DIAMOND, in pp. Means and 95\% confidence intervals use 10,000 bootstrap draws over independent sources. Intervals are conditional on the selected editor and do not include training-seed variation. Endpoint is the paired semantic gain at $h=8$.}\label{tab:confirmation-diamond}
\begin{tabular}{@{}llrlrlrl@{}}
\toprule
Task & Tool & \multicolumn{2}{c}{ISG} & \multicolumn{2}{c}{SSG@8} & \multicolumn{2}{c}{Endpoint}\\
\cmidrule(lr){3-4}\cmidrule(lr){5-6}\cmidrule(lr){7-8}
& & Mean & 95\% CI & Mean & 95\% CI & Mean & 95\% CI\\
\midrule
\multirow{12}{*}{Crafter} & Mean training delta & +0.11 & [-0.10, 0.37] & +0.31 & [0.04, 0.61] & +0.51 & [0.10, 0.96]\\
 & Inverse probe & +0.89 & [-1.75, 4.31] & +0.21 & [-1.54, 2.46] & -0.71 & [-2.85, 1.49]\\
 & Normalized probe & -0.23 & [-1.66, 1.12] & -1.06 & [-2.81, 0.43] & -2.37 & [-5.50, -0.03]\\
 & Difference of means & -0.43 & [-1.11, 0.08] & +0.09 & [-0.14, 0.34] & +0.42 & [0.08, 0.80]\\
 & PCA + ridge & -0.16 & [-0.39, 0.02] & -0.03 & [-0.20, 0.15] & -0.00075 & [-0.17, 0.20]\\
 & Uncentered carrier & -0.15 & [-0.38, 0.02] & -0.01 & [-0.15, 0.17] & -0.02 & [-0.17, 0.15]\\
 & Immediate LoReFT & -0.50 & [-1.12, -0.0019] & +0.18 & [-0.30, 0.66] & +0.47 & [-0.05, 1.04]\\
 & Delayed LoReFT & -0.27 & [-0.80, 0.17] & +0.24 & [-0.36, 0.81] & +0.76 & [-0.04, 1.63]\\
 & Nonlinear MLP & -0.07 & [-0.23, +0.03] & +0.08 & [-0.05, +0.26] & +0.22 & [+0.07, +0.41]\\
 & Manifold--WM & -0.20 & [-0.76, +0.19] & -0.11 & [-0.32, +0.06] & +0.06 & [-0.23, +0.32]\\
 & Concept DAS--WM & -0.08 & [-0.38, +0.21] & -0.07 & [-0.24, +0.06] & -0.00031 & [-0.21, +0.25]\\
 & RAP & +15.23 & [9.83, 21.61] & +10.76 & [6.65, 15.74] & +8.02 & [4.57, 11.98]\\
\addlinespace
\multirow{12}{*}{Cartpole} & Mean training delta & +0.09 & [-0.05, 0.27] & +0.06 & [-0.01, 0.13] & +0.17 & [-0.01, 0.38]\\
 & Inverse probe & +0.16 & [-0.07, 0.41] & -0.13 & [-0.38, 0.12] & +0.13 & [-0.27, 0.54]\\
 & Normalized probe & +0.01 & [0.00, 0.02] & +0.02 & [-0.0029, 0.04] & +0.07 & [-0.02, 0.17]\\
 & Difference of means & +0.00039 & [-0.02, 0.02] & +0.0040 & [-0.03, 0.03] & +0.04 & [-0.07, 0.14]\\
 & PCA + ridge & -0.06 & [-0.32, 0.17] & +0.24 & [0.03, 0.43] & +0.44 & [0.10, 0.84]\\
 & Uncentered carrier & -0.01 & [-0.09, 0.08] & +0.04 & [0.01, 0.07] & +0.10 & [-0.01, 0.21]\\
 & Immediate LoReFT & +0.02 & [-0.17, 0.23] & +0.32 & [0.09, 0.58] & +0.45 & [-0.14, 1.14]\\
 & Delayed LoReFT & +0.34 & [-0.02, 0.72] & +3.24 & [1.87, 4.61] & +4.92 & [2.95, 6.95]\\
 & Nonlinear MLP & +0.04 & [-0.12, +0.22] & +0.07 & [-0.02, +0.17] & +0.15 & [-0.14, +0.50]\\
 & Manifold--WM & -0.67 & [-1.46, +0.12] & -2.94 & [-4.67, -1.20] & -3.14 & [-5.54, -0.79]\\
 & Concept DAS--WM & -0.18 & [-0.67, +0.29] & -0.14 & [-0.78, +0.49] & +0.03 & [-0.98, +1.03]\\
 & RAP & +5.97 & [4.84, 7.14] & +9.01 & [7.19, 10.71] & +12.71 & [9.86, 15.43]\\
\addlinespace
\multirow{12}{*}{CoinRun} & Mean training delta & +0.06 & [0.04, 0.08] & +0.14 & [0.03, 0.25] & +0.24 & [-0.04, 0.57]\\
 & Inverse probe & +0.01 & [0.00036, 0.01] & +0.02 & [-0.01, 0.05] & +0.05 & [-0.02, 0.13]\\
 & Normalized probe & +3.05 & [2.07, 4.18] & +2.25 & [0.78, 3.76] & +0.84 & [-1.26, 2.99]\\
 & Difference of means & +0.34 & [0.03, 0.65] & +0.16 & [-0.53, 0.86] & -0.20 & [-1.18, 0.82]\\
 & PCA + ridge & +0.88 & [0.30, 1.52] & +0.20 & [-0.68, 1.16] & -0.63 & [-1.83, 0.60]\\
 & Uncentered carrier & +0.81 & [0.26, 1.45] & +0.27 & [-0.58, 1.24] & -0.55 & [-1.74, 0.70]\\
 & Immediate LoReFT & +0.66 & [0.01, 1.46] & +0.39 & [-0.77, 1.60] & +0.37 & [-0.98, 1.78]\\
 & Delayed LoReFT & -8.67 & [-9.55, -7.79] & -1.17 & [-3.09, 0.69] & -1.80 & [-4.57, 0.87]\\
 & Nonlinear MLP & +0.04 & [+0.03, +0.06] & +0.09 & [-0.02, +0.19] & +0.17 & [-0.09, +0.48]\\
 & Manifold--WM & +0.48 & [+0.17, +0.89] & +0.34 & [-0.47, +1.23] & -0.11 & [-1.31, +1.17]\\
 & Concept DAS--WM & +0.40 & [+0.08, +0.88] & -0.23 & [-0.73, +0.26] & -0.76 & [-1.64, +0.01]\\
 & RAP & +7.62 & [6.06, 9.25] & +4.14 & [2.20, 6.07] & +1.47 & [-0.83, 3.76]\\
\bottomrule
\end{tabular}
\end{table}

%% file: tables/combined_statistics_dreamerv3.tex
\begin{table}[!htbp]
\centering\scriptsize\setlength{\tabcolsep}{2.2pt}
\caption{Complete intervention results for DreamerV3, in pp. Means and 95\% confidence intervals use 10,000 bootstrap draws over independent sources. Intervals are conditional on the selected editor and do not include training-seed variation. Endpoint is the paired semantic gain at $h=8$.}\label{tab:confirmation-dreamerv3}
\begin{tabular}{@{}llrlrlrl@{}}
\toprule
Task & Tool & \multicolumn{2}{c}{ISG} & \multicolumn{2}{c}{SSG@8} & \multicolumn{2}{c}{Endpoint}\\
\cmidrule(lr){3-4}\cmidrule(lr){5-6}\cmidrule(lr){7-8}
& & Mean & 95\% CI & Mean & 95\% CI & Mean & 95\% CI\\
\midrule
\multirow{12}{*}{Crafter} & Mean training delta & +0.11 & [-0.38, 0.62] & -0.13 & [-0.41, 0.14] & +0.06 & [-0.31, 0.47]\\
 & Inverse probe & +0.28 & [-0.25, 0.84] & +0.19 & [-0.02, 0.38] & -0.07 & [-0.56, 0.39]\\
 & Normalized probe & -0.01 & [-0.59, 0.54] & +0.32 & [-0.12, 0.76] & +0.26 & [-0.48, 0.95]\\
 & Difference of means & +0.06 & [-0.15, 0.29] & -0.04 & [-0.18, 0.08] & +0.08 & [-0.24, 0.38]\\
 & PCA + ridge & +0.11 & [-0.44, 0.68] & -0.15 & [-0.48, 0.17] & -0.20 & [-0.80, 0.36]\\
 & Uncentered carrier & +0.12 & [-0.25, 0.48] & -0.10 & [-0.30, 0.11] & +0.10 & [-0.28, 0.47]\\
 & Immediate LoReFT & -0.21 & [-0.73, 0.28] & -0.12 & [-0.41, 0.15] & +0.10 & [-0.47, 0.67]\\
 & Delayed LoReFT & -0.06 & [-0.43, 0.28] & -0.17 & [-0.33, -0.03] & -0.19 & [-0.69, 0.26]\\
 & Nonlinear MLP & +0.05 & [-0.16, +0.26] & -0.02 & [-0.19, +0.11] & +0.15 & [-0.08, +0.37]\\
 & Manifold--WM & +0.03 & [-0.37, +0.43] & -0.04 & [-0.36, +0.31] & -0.16 & [-0.70, +0.37]\\
 & Concept DAS--WM & +0.30 & [-0.15, +0.78] & +0.05 & [-0.11, +0.22] & -0.03 & [-0.45, +0.37]\\
 & RAP & +2.71 & [1.74, 3.81] & +1.24 & [0.75, 1.78] & +0.74 & [0.01, 1.51]\\
\addlinespace
\multirow{12}{*}{Cartpole} & Mean training delta & +0.13 & [-0.13, 0.42] & +0.10 & [-0.14, 0.34] & +0.22 & [-0.14, 0.60]\\
 & Inverse probe & -0.16 & [-0.32, -0.03] & -0.03 & [-0.12, 0.05] & +0.02 & [-0.12, 0.15]\\
 & Normalized probe & +0.19 & [-0.30, 0.67] & +0.54 & [0.19, 0.88] & +0.80 & [0.34, 1.26]\\
 & Difference of means & -0.35 & [-0.97, 0.24] & +0.03 & [-0.38, 0.44] & -0.25 & [-0.94, 0.39]\\
 & PCA + ridge & +0.88 & [0.60, 1.17] & +1.02 & [0.75, 1.29] & +1.01 & [0.57, 1.50]\\
 & Uncentered carrier & +0.74 & [0.40, 1.07] & +1.09 & [0.81, 1.38] & +1.12 & [0.63, 1.63]\\
 & Immediate LoReFT & +2.57 & [2.01, 3.15] & +3.22 & [2.81, 3.63] & +3.91 & [3.24, 4.63]\\
 & Delayed LoReFT & +2.39 & [1.79, 2.98] & +3.46 & [3.01, 3.94] & +4.07 & [3.31, 4.86]\\
 & Nonlinear MLP & +0.70 & [+0.41, +1.01] & +0.76 & [+0.51, +1.02] & +0.83 & [+0.39, +1.29]\\
 & Manifold--WM & +0.62 & [+0.12, +1.13] & +0.60 & [+0.12, +1.10] & +0.44 & [-0.17, +1.09]\\
 & Concept DAS--WM & +0.40 & [-0.0050, +0.79] & +0.44 & [+0.11, +0.76] & +0.43 & [-0.02, +0.89]\\
 & RAP & +0.52 & [0.29, 0.78] & +0.92 & [0.64, 1.20] & +0.91 & [0.50, 1.34]\\
\addlinespace
\multirow{12}{*}{CoinRun} & Mean training delta & +0.0044 & [-0.0020, 0.01] & +0.0017 & [-0.0040, 0.01] & -0.0046 & [-0.01, 0.0039]\\
 & Inverse probe & +0.03 & [-0.07, 0.13] & -0.07 & [-0.20, 0.07] & -0.14 & [-0.30, 0.03]\\
 & Normalized probe & +0.18 & [0.08, 0.29] & +0.18 & [0.03, 0.34] & +0.13 & [-0.05, 0.33]\\
 & Difference of means & +0.0023 & [-0.01, 0.01] & +0.01 & [-0.0030, 0.03] & -0.0011 & [-0.02, 0.02]\\
 & PCA + ridge & +0.10 & [-0.09, 0.29] & +0.37 & [0.01, 0.81] & +0.42 & [-0.03, 0.95]\\
 & Uncentered carrier & +0.09 & [-0.10, 0.28] & +0.38 & [0.02, 0.84] & +0.47 & [0.03, 1.01]\\
 & Immediate LoReFT & +0.07 & [-0.07, 0.21] & +0.30 & [0.07, 0.62] & +0.19 & [-0.07, 0.51]\\
 & Delayed LoReFT & +0.06 & [-0.08, 0.19] & +0.14 & [-0.05, 0.33] & +0.14 & [-0.12, 0.39]\\
 & Nonlinear MLP & +0.00096 & [-0.01, +0.01] & +0.000055 & [-0.01, +0.01] & +0.0018 & [-0.02, +0.03]\\
 & Manifold--WM & +0.02 & [+0.000022, +0.03] & +0.01 & [-0.01, +0.03] & -0.02 & [-0.07, +0.03]\\
 & Concept DAS--WM & +0.06 & [-0.12, +0.23] & +0.31 & [+0.0044, +0.65] & +0.37 & [+0.01, +0.76]\\
 & RAP & +1.06 & [0.78, 1.38] & +1.20 & [0.76, 1.77] & +1.18 & [0.64, 1.91]\\
\bottomrule
\end{tabular}
\end{table}

%% file: tables/combined_statistics_storm.tex
\begin{table}[!htbp]
\centering\scriptsize\setlength{\tabcolsep}{2.2pt}
\caption{Complete intervention results for STORM, in pp. Means and 95\% confidence intervals use 10,000 bootstrap draws over independent sources. Intervals are conditional on the selected editor and do not include training-seed variation. Endpoint is the paired semantic gain at $h=8$.}\label{tab:confirmation-storm}
\begin{tabular}{@{}llrlrlrl@{}}
\toprule
Task & Tool & \multicolumn{2}{c}{ISG} & \multicolumn{2}{c}{SSG@8} & \multicolumn{2}{c}{Endpoint}\\
\cmidrule(lr){3-4}\cmidrule(lr){5-6}\cmidrule(lr){7-8}
& & Mean & 95\% CI & Mean & 95\% CI & Mean & 95\% CI\\
\midrule
\multirow{12}{*}{Crafter} & Mean training delta & -0.87 & [-1.65, -0.13] & -0.03 & [-0.47, 0.44] & -0.28 & [-1.13, 0.56]\\
 & Inverse probe & +2.17 & [0.63, 3.70] & +2.06 & [1.07, 3.12] & +0.85 & [-0.24, 1.92]\\
 & Normalized probe & +0.62 & [-0.87, 2.05] & +1.44 & [0.53, 2.34] & +0.11 & [-1.26, 1.39]\\
 & Difference of means & +0.24 & [-0.37, 0.81] & +0.09 & [-0.21, 0.38] & -0.08 & [-0.86, 0.72]\\
 & PCA + ridge & -0.33 & [-1.19, 0.59] & +0.60 & [-0.01, 1.33] & +0.63 & [-0.51, 1.89]\\
 & Uncentered carrier & -0.17 & [-0.86, 0.61] & +0.17 & [-0.27, 0.74] & +0.25 & [-0.65, 1.19]\\
 & Immediate LoReFT & -1.27 & [-2.32, -0.41] & -1.07 & [-2.26, -0.09] & -1.39 & [-2.98, 0.02]\\
 & Delayed LoReFT & -3.73 & [-5.71, -2.08] & -0.60 & [-1.51, 0.29] & -1.18 & [-2.68, 0.14]\\
 & Nonlinear MLP & -0.66 & [-1.44, +0.09] & +0.22 & [-0.22, +0.68] & -0.12 & [-1.05, +0.80]\\
 & Manifold--WM & +0.13 & [-0.21, +0.53] & +0.24 & [+0.07, +0.42] & +0.37 & [-0.23, +0.96]\\
 & Concept DAS--WM & -0.04 & [-0.72, +0.61] & +0.09 & [-0.34, +0.53] & -0.78 & [-1.61, +0.01]\\
 & RAP & +8.32 & [5.38, 11.68] & +3.62 & [2.09, 5.48] & +1.97 & [0.33, 3.65]\\
\addlinespace
\multirow{12}{*}{Cartpole} & Mean training delta & +0.22 & [-0.11, 0.56] & +0.37 & [0.02, 0.74] & +0.30 & [-0.17, 0.81]\\
 & Inverse probe & -0.28 & [-0.57, -0.02] & +0.02 & [-0.30, 0.31] & -0.18 & [-0.60, 0.21]\\
 & Normalized probe & -0.03 & [-0.44, 0.37] & +0.21 & [-0.10, 0.52] & +0.26 & [-0.20, 0.72]\\
 & Difference of means & -0.13 & [-0.29, -0.00068] & -0.07 & [-0.34, 0.19] & -0.24 & [-0.64, 0.12]\\
 & PCA + ridge & +0.07 & [-0.32, 0.45] & +0.07 & [-0.25, 0.37] & +0.04 & [-0.33, 0.40]\\
 & Uncentered carrier & +0.0048 & [-0.34, 0.32] & +0.03 & [-0.27, 0.32] & -0.04 & [-0.46, 0.36]\\
 & Immediate LoReFT & +0.82 & [-0.18, 1.82] & +0.59 & [0.01, 1.17] & +0.47 & [-0.35, 1.32]\\
 & Delayed LoReFT & -1.21 & [-2.88, 0.45] & +0.75 & [-0.21, 1.71] & +0.73 & [-0.46, 1.99]\\
 & Nonlinear MLP & +0.85 & [+0.28, +1.42] & +0.66 & [+0.26, +1.04] & +0.75 & [+0.23, +1.25]\\
 & Manifold--WM & -2.78 & [-5.04, -0.67] & -1.16 & [-2.66, +0.17] & -1.40 & [-2.95, +0.01]\\
 & Concept DAS--WM & +0.08 & [-0.11, +0.27] & +0.05 & [-0.18, +0.26] & +0.02 & [-0.25, +0.31]\\
 & RAP & +12.05 & [10.44, 13.58] & +8.90 & [8.01, 9.80] & +8.09 & [6.83, 9.38]\\
\addlinespace
\multirow{12}{*}{CoinRun} & Mean training delta & -0.01 & [-0.05, 0.02] & -0.04 & [-0.10, -0.01] & -0.06 & [-0.19, 0.04]\\
 & Inverse probe & +0.03 & [-0.21, 0.26] & -0.05 & [-0.40, 0.21] & -0.17 & [-0.77, 0.26]\\
 & Normalized probe & +0.71 & [0.28, 1.11] & +0.44 & [0.16, 0.74] & +0.18 & [-0.36, 0.72]\\
 & Difference of means & +0.05 & [-0.01, 0.10] & -0.02 & [-0.12, 0.07] & -0.02 & [-0.22, 0.18]\\
 & PCA + ridge & -0.0038 & [-0.01, 0.0039] & -0.01 & [-0.08, 0.08] & -0.10 & [-0.25, 0.02]\\
 & Uncentered carrier & -0.0040 & [-0.01, 0.0035] & -0.01 & [-0.08, 0.07] & -0.10 & [-0.25, 0.02]\\
 & Immediate LoReFT & +0.03 & [-0.05, 0.11] & +0.08 & [-0.05, 0.23] & +0.12 & [-0.11, 0.40]\\
 & Delayed LoReFT & -1.51 & [-1.97, -1.03] & -1.06 & [-1.70, -0.51] & -1.43 & [-2.43, -0.55]\\
 & Nonlinear MLP & +0.01 & [-0.01, +0.03] & -0.05 & [-0.11, -0.01] & -0.13 & [-0.32, +0.02]\\
 & Manifold--WM & +0.06 & [-0.07, +0.20] & +0.04 & [-0.15, +0.21] & -0.11 & [-0.46, +0.18]\\
 & Concept DAS--WM & -0.01 & [-0.04, +0.02] & -0.03 & [-0.11, +0.06] & -0.15 & [-0.33, -0.01]\\
 & RAP & +6.17 & [4.63, 7.98] & +3.35 & [2.24, 4.61] & +2.65 & [1.43, 3.97]\\
\bottomrule
\end{tabular}
\end{table}

%% file: tables/scoreable_event_coverage.tex
\begin{table}[!htbp]
\centering\scriptsize\setlength{\tabcolsep}{1.7pt}
\caption{Scoreable event coverage at every evaluated horizon. Entries from $h=0$ through $h=8$ are the number of scoreable events out of the total in the Events column. The mask is determined from real references and is shared across all models, interventions, and three prediction seeds.}
\label{tab:scoreable-event-coverage}
\begin{tabular}{@{}lrrrrrrrrrrr@{}}
\toprule
Task & Sources & Events & $h=0$ & $h=1$ & $h=2$ & $h=3$ & $h=4$ & $h=5$ & $h=6$ & $h=7$ & $h=8$\\
\midrule
Crafter & 38 & 41 & 41 & 40 & 41 & 41 & 41 & 40 & 41 & 41 & 41\\
Cartpole & 48 & 96 & 96 & 96 & 96 & 96 & 96 & 96 & 96 & 96 & 96\\
CoinRun & 47 & 92 & 92 & 92 & 92 & 92 & 92 & 92 & 92 & 92 & 92\\
\bottomrule
\end{tabular}
\end{table}

%% file: figures/special_cases.tex
\begin{figure}[!htbp]
\centering\includegraphics[width=\linewidth]{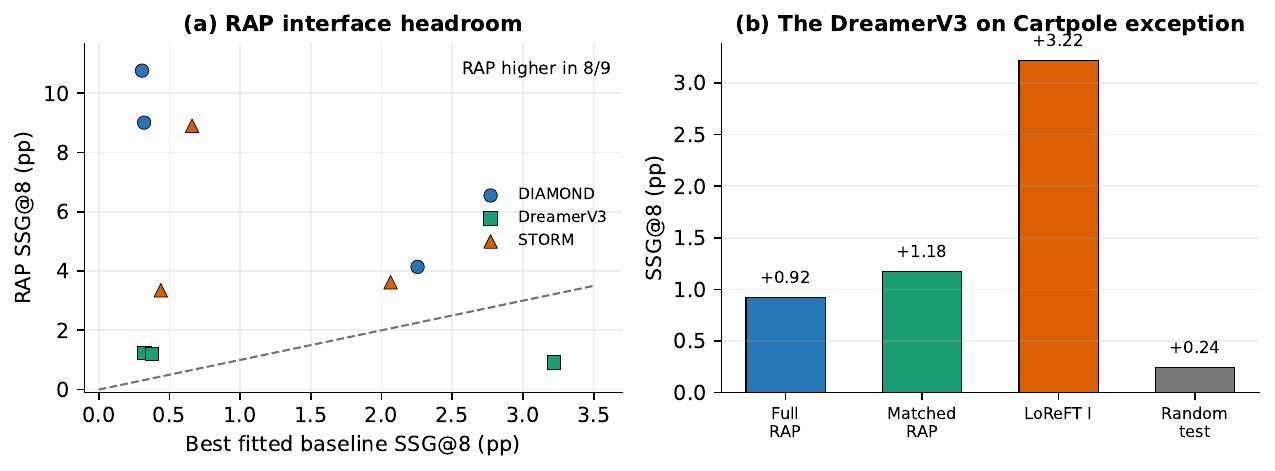}
\caption{\textbf{The dominant interface pattern and its exception.} (a) RAP exceeds the strongest fitted editor in eight of nine cohorts spanning models and tasks. (b) DreamerV3 on Cartpole is the exception: Immediate LoReFT exceeds full RAP, RAP matched by norm, and its random test edit matched by norm. Delayed LoReFT is excluded from both panels.}\label{fig:special-cases}
\end{figure}

%% file: tables/special_case_robustness_v5.tex
\begin{table}[!htbp]
\centering\small
\caption{DreamerV3 on Cartpole exception across the three registered prediction seeds. Values are SSG@8 in pp with equal source weights. The paired 95\% bootstrap interval over sources and all seeds for RAP minus Immediate LoReFT is [-2.68, -1.90]. Seeds are robustness strata over the same 48 sources, not independent datasets.}
\label{tab:dreamer-exception-seeds}
\begin{tabular}{@{}lrrr@{}}\toprule
Prediction seed & RAP & LoReFT I & RAP minus LoReFT I\\\midrule
69000 & +0.76 & +3.39 & -2.63\\
69001 & +1.18 & +3.70 & -2.53\\
69002 & +0.84 & +2.56 & -1.73\\
All seeds & +0.92 & +3.22 & -2.30\\
\bottomrule\end{tabular}
\end{table}

%% file: tables/paired_ranking_ci.tex
\begin{table}[!htbp]
\centering\scriptsize
\setlength{\tabcolsep}{3pt}
\caption{The fitted candidate set includes all eleven methods (the ten baseline editors and Delayed LoReFT). Exploratory paired bootstrap intervals over sources for observed SSG@8 leaders. Seeds and events are averaged within source before 10,000 resamples. The final column uses $\alpha=0.05/9$ for the nine cohort contrasts in each contrast family. Because leaders were selected by the same point estimates, these intervals describe ranking stability rather than preregistered hypothesis tests.}\label{tab:paired-ranking-ci}
\begin{tabular}{@{}lllrrr@{}}\toprule
WM & Task & Paired SSG@8 contrast & Mean & Pointwise 95\% CI & Simultaneous 99.44\% CI\\\midrule
DIAMOND & Crafter & RAP--Mean & +10.46 & [+6.34, +15.40] & [+4.98, +17.82]\\
DIAMOND & Crafter & Mean--Delayed & +0.06 & [-0.31, +0.53] & [-0.45, +0.75]\\
DIAMOND & Cartpole & RAP--Delayed & +5.77 & [+4.21, +7.33] & [+3.60, +7.94]\\
DIAMOND & Cartpole & Delayed--Immediate & +2.92 & [+1.64, +4.19] & [+1.13, +4.78]\\
DIAMOND & CoinRun & RAP--NormProbe & +1.88 & [+0.65, +3.17] & [+0.19, +3.74]\\
DIAMOND & CoinRun & NormProbe--Immediate & +1.86 & [+0.70, +3.10] & [+0.23, +3.59]\\
DreamerV3 & Crafter & RAP--NormProbe & +0.91 & [+0.49, +1.37] & [+0.32, +1.55]\\
DreamerV3 & Crafter & NormProbe--InvProbe & +0.14 & [-0.24, +0.56] & [-0.40, +0.76]\\
DreamerV3 & Cartpole & RAP--Delayed & -2.54 & [-2.97, -2.10] & [-3.14, -1.93]\\
DreamerV3 & Cartpole & Delayed--Immediate & +0.24 & [+0.08, +0.41] & [+0.01, +0.47]\\
DreamerV3 & CoinRun & RAP--Carrier & +0.82 & [+0.54, +1.11] & [+0.45, +1.23]\\
DreamerV3 & CoinRun & Carrier--PCA+ridge & +0.01 & [-0.02, +0.04] & [-0.03, +0.05]\\
STORM & Crafter & RAP--InvProbe & +1.56 & [+0.28, +2.95] & [-0.18, +3.59]\\
STORM & Crafter & InvProbe--NormProbe & +0.62 & [-0.04, +1.35] & [-0.28, +1.66]\\
STORM & Cartpole & RAP--Delayed & +8.14 & [+7.09, +9.23] & [+6.63, +9.71]\\
STORM & Cartpole & Delayed--Nonlinear MLP & +0.09 & [-0.94, +1.08] & [-1.34, +1.50]\\
STORM & CoinRun & RAP--NormProbe & +2.91 & [+1.93, +4.03] & [+1.63, +4.52]\\
STORM & CoinRun & NormProbe--Immediate & +0.36 & [+0.05, +0.67] & [-0.07, +0.80]\\
\bottomrule\end{tabular}
\end{table}

%% file: tables/intervention_diagnostics.tex
\begin{table}[t]
\centering\scriptsize\setlength{\tabcolsep}{3pt}
\caption{Intervention diagnostics for the two highest and lowest SSG@8 tools from the original nine-intervention bank in each available model and task. Ties follow the fixed method order. Norm is the mean Euclidean edit norm with equal source weights, concatenating the three edits across denoiser calls for DIAMOND; units differ across interfaces. $\Delta$ random is SSG@8 above the random control matched by norm. Both gain columns are in pp. All original primary tools produce nonzero updates on every evaluated pair of events and seeds. SSG@8 for requests with zero change is evaluated for the eight fitted tools with targets equal to the descriptor of the predicted image; N/A marks RAP. Complete diagnostics for every evaluated tool accompany the results bound to each source.}\label{tab:intervention-diagnostics}
\begin{tabular}{@{}lllrrr@{}}\toprule
WM & Task & Tool & Norm & $\Delta$ random & SSG@8 for zero request\\\midrule
DIAMOND & Crafter & Mean training delta & 40.58 & +0.27 & +0.31\\
DIAMOND & Crafter & Normalized probe & 114.76 & -0.83 & -2.16\\
DIAMOND & Crafter & RAP & 107.43 & +10.89 & N/A\\
DIAMOND & Cartpole & Inverse probe & 14.06 & -0.14 & -0.20\\
DIAMOND & Cartpole & LoReFT delayed & 37.47 & +3.27 & +3.24\\
DIAMOND & Cartpole & RAP & 23.90 & +9.06 & N/A\\
DIAMOND & CoinRun & Normalized probe & 290.14 & +2.25 & -1.11\\
DIAMOND & CoinRun & LoReFT delayed & 580.29 & -1.16 & -1.17\\
DIAMOND & CoinRun & RAP & 272.40 & +4.10 & N/A\\
DreamerV3 & Crafter & Normalized probe & 23.13 & +0.39 & +0.12\\
DreamerV3 & Crafter & LoReFT delayed & 5.94 & -0.12 & -0.17\\
DreamerV3 & Crafter & RAP & 73.64 & +1.37 & N/A\\
DreamerV3 & Cartpole & Inverse probe & 1.17 & +0.03 & -0.05\\
DreamerV3 & Cartpole & LoReFT immediate & 21.88 & +2.98 & +3.22\\
DreamerV3 & Cartpole & LoReFT delayed & 21.88 & +3.36 & +3.46\\
DreamerV3 & CoinRun & Inverse probe & 12.94 & -0.07 & -0.02\\
DreamerV3 & CoinRun & Uncentered carrier & 25.76 & +0.44 & +0.08\\
DreamerV3 & CoinRun & RAP & 31.10 & +1.21 & N/A\\
STORM & Crafter & Inverse probe & 22.40 & +2.27 & +0.31\\
STORM & Crafter & LoReFT immediate & 46.68 & -1.38 & -1.08\\
STORM & Crafter & RAP & 33.17 & +3.65 & N/A\\
STORM & Cartpole & Difference of means & 0.63 & +0.02 & -0.05\\
STORM & Cartpole & LoReFT delayed & 20.92 & +0.91 & +0.75\\
STORM & Cartpole & RAP & 14.63 & +8.98 & N/A\\
STORM & CoinRun & Normalized probe & 26.65 & +0.62 & -0.03\\
STORM & CoinRun & LoReFT delayed & 53.30 & -1.05 & -1.06\\
STORM & CoinRun & RAP & 35.27 & +3.53 & N/A\\
\bottomrule\end{tabular}
\end{table}

%% file: tables/random_controls_panel_diamond.tex
\begin{table}[!htbp]
\centering\scriptsize\setlength{\tabcolsep}{1.7pt}
\caption{Validation, test and norm matched random controls for DIAMOND. $\Delta$ is the test editor effect minus its random control.}\label{tab:all-random-controls-panel-diamond}
\begin{tabular}{@{}lrrrrrrrrrrrr@{}}
\toprule
\multicolumn{13}{c}{ISG (pp)}\\
& \multicolumn{4}{c}{Crafter} & \multicolumn{4}{c}{Cartpole} & \multicolumn{4}{c}{CoinRun}\\
\cmidrule(lr){2-5}\cmidrule(lr){6-9}\cmidrule(lr){10-13}
Tool & Val. & Test & Rand. & $\Delta$ & Val. & Test & Rand. & $\Delta$ & Val. & Test & Rand. & $\Delta$\\
\midrule
Mean training delta & -0.23 & +0.11 & -0.03 & +0.14 & +0.88 & +0.09 & +0.06 & +0.03 & +0.06 & +0.06 & +0.00 & +0.06\\
Inverse probe & +0.99 & +0.89 & -0.30 & +1.19 & +0.98 & +0.16 & +0.09 & +0.08 & -0.01 & +0.01 & +0.00 & +0.01\\
Normalized probe & -1.13 & -0.23 & -0.15 & -0.08 & +0.00 & +0.01 & +0.00 & +0.01 & +3.80 & +3.05 & -0.00 & +3.05\\
Difference of means & -0.84 & -0.43 & -0.00 & -0.43 & +0.00 & +0.00 & +0.00 & +0.00 & +0.18 & +0.34 & -0.00 & +0.34\\
PCA + ridge & -0.30 & -0.16 & +0.01 & -0.16 & +0.88 & -0.06 & +0.03 & -0.09 & +1.17 & +0.88 & -0.00 & +0.88\\
Uncentered carrier & -0.30 & -0.15 & -0.02 & -0.13 & +0.78 & -0.01 & -0.00 & -0.00 & +1.33 & +0.81 & -0.01 & +0.82\\
Immediate LoReFT & -1.64 & -0.50 & +0.02 & -0.52 & +0.66 & +0.02 & +0.03 & -0.02 & +2.29 & +0.66 & -0.03 & +0.69\\
Delayed LoReFT & -0.88 & -0.27 & +0.02 & -0.29 & +0.58 & +0.34 & +0.16 & +0.18 & -8.81 & -8.67 & -0.03 & -8.64\\
Nonlinear MLP & -0.06 & -0.07 & +0.04 & -0.11 & +0.88 & +0.04 & +0.08 & -0.04 & +0.05 & +0.04 & +0.00094 & +0.04\\
Manifold--WM & -0.38 & -0.20 & +0.02 & -0.22 & +1.69 & -0.67 & +0.10 & -0.77 & +0.39 & +0.48 & +0.0015 & +0.48\\
Concept DAS--WM & -0.18 & -0.08 & -0.01 & -0.07 & +1.48 & -0.18 & +0.0042 & -0.19 & +0.74 & +0.40 & +0.00064 & +0.40\\
\addlinespace[2pt]
\multicolumn{13}{c}{SSG@8 (pp)}\\
& \multicolumn{4}{c}{Crafter} & \multicolumn{4}{c}{Cartpole} & \multicolumn{4}{c}{CoinRun}\\
\cmidrule(lr){2-5}\cmidrule(lr){6-9}\cmidrule(lr){10-13}
Tool & Val. & Test & Rand. & $\Delta$ & Val. & Test & Rand. & $\Delta$ & Val. & Test & Rand. & $\Delta$\\
\midrule
Mean training delta & +0.94 & +0.31 & +0.04 & +0.27 & +0.19 & +0.06 & -0.00 & +0.06 & +0.06 & +0.14 & -0.01 & +0.15\\
Inverse probe & +2.71 & +0.21 & -0.35 & +0.56 & +0.96 & -0.13 & +0.01 & -0.14 & +0.02 & +0.02 & +0.00 & +0.02\\
Normalized probe & +4.63 & -1.06 & -0.23 & -0.83 & -0.02 & +0.02 & -0.00 & +0.02 & +3.17 & +2.25 & +0.01 & +2.25\\
Difference of means & +0.91 & +0.09 & -0.00 & +0.10 & -0.01 & +0.00 & -0.01 & +0.01 & +0.71 & +0.16 & -0.04 & +0.20\\
PCA + ridge & +0.31 & -0.03 & -0.03 & +0.00 & +0.15 & +0.24 & -0.04 & +0.27 & +0.61 & +0.20 & +0.01 & +0.19\\
Uncentered carrier & +0.24 & -0.01 & -0.01 & +0.00 & +0.02 & +0.04 & -0.00 & +0.04 & +1.01 & +0.27 & -0.01 & +0.27\\
Immediate LoReFT & +0.99 & +0.18 & +0.03 & +0.16 & +0.89 & +0.32 & +0.01 & +0.31 & +2.31 & +0.39 & -0.04 & +0.44\\
Delayed LoReFT & +1.41 & +0.24 & -0.00 & +0.25 & +8.56 & +3.24 & -0.03 & +3.27 & +1.06 & -1.17 & -0.01 & -1.16\\
Nonlinear MLP & +0.86 & +0.08 & +0.01 & +0.08 & +0.23 & +0.07 & -0.02 & +0.09 & +0.07 & +0.09 & +0.01 & +0.08\\
Manifold--WM & +0.53 & -0.11 & -0.05 & -0.06 & +3.83 & -2.94 & +0.0030 & -2.95 & +0.32 & +0.34 & +0.02 & +0.32\\
Concept DAS--WM & +0.54 & -0.07 & +0.01 & -0.08 & +2.32 & -0.14 & +0.02 & -0.17 & +0.61 & -0.23 & -0.01 & -0.21\\
\bottomrule
\end{tabular}
\end{table}

%% file: tables/random_controls_panel_dreamerv3.tex
\begin{table}[!htbp]
\centering\scriptsize\setlength{\tabcolsep}{1.7pt}
\caption{Validation, test and norm matched random controls for DreamerV3. $\Delta$ is the test editor effect minus its random control.}\label{tab:all-random-controls-panel-dreamerv3}
\begin{tabular}{@{}lrrrrrrrrrrrr@{}}
\toprule
\multicolumn{13}{c}{ISG (pp)}\\
& \multicolumn{4}{c}{Crafter} & \multicolumn{4}{c}{Cartpole} & \multicolumn{4}{c}{CoinRun}\\
\cmidrule(lr){2-5}\cmidrule(lr){6-9}\cmidrule(lr){10-13}
Tool & Val. & Test & Rand. & $\Delta$ & Val. & Test & Rand. & $\Delta$ & Val. & Test & Rand. & $\Delta$\\
\midrule
Mean training delta & +0.11 & +0.11 & -0.06 & +0.17 & +0.57 & +0.13 & -0.05 & +0.18 & +0.00 & +0.00 & -0.00 & +0.00\\
Inverse probe & +0.59 & +0.28 & +0.04 & +0.23 & +0.00 & -0.16 & -0.09 & -0.07 & +0.18 & +0.03 & -0.03 & +0.06\\
Normalized probe & +0.14 & -0.01 & +0.41 & -0.41 & -0.55 & +0.19 & -0.14 & +0.33 & +0.24 & +0.18 & +0.04 & +0.14\\
Difference of means & -0.16 & +0.06 & -0.14 & +0.20 & -0.38 & -0.35 & -0.12 & -0.24 & -0.00 & +0.00 & +0.00 & -0.00\\
PCA + ridge & -0.09 & +0.11 & +0.19 & -0.09 & +0.44 & +0.88 & -0.02 & +0.90 & +0.08 & +0.10 & +0.02 & +0.08\\
Uncentered carrier & -0.57 & +0.12 & +0.46 & -0.33 & +1.13 & +0.74 & -0.13 & +0.87 & +0.13 & +0.09 & -0.03 & +0.12\\
Immediate LoReFT & -0.49 & -0.21 & -0.11 & -0.10 & +2.39 & +2.57 & +0.01 & +2.55 & +0.41 & +0.07 & +0.02 & +0.05\\
Delayed LoReFT & -0.05 & -0.06 & +0.04 & -0.10 & +2.93 & +2.39 & -0.01 & +2.40 & +0.38 & +0.06 & +0.08 & -0.02\\
Nonlinear MLP & -0.14 & +0.05 & -0.08 & +0.13 & +0.49 & +0.70 & -0.05 & +0.75 & +0.0015 & +0.00096 & +0.01 & -0.0050\\
Manifold--WM & +0.62 & +0.03 & +0.37 & -0.34 & +1.95 & +0.62 & -0.14 & +0.76 & +0.02 & +0.02 & -0.00067 & +0.02\\
Concept DAS--WM & +0.02 & +0.30 & -0.03 & +0.33 & -0.07 & +0.40 & -0.22 & +0.62 & -0.08 & +0.06 & +0.04 & +0.02\\
\addlinespace[2pt]
\multicolumn{13}{c}{SSG@8 (pp)}\\
& \multicolumn{4}{c}{Crafter} & \multicolumn{4}{c}{Cartpole} & \multicolumn{4}{c}{CoinRun}\\
\cmidrule(lr){2-5}\cmidrule(lr){6-9}\cmidrule(lr){10-13}
Tool & Val. & Test & Rand. & $\Delta$ & Val. & Test & Rand. & $\Delta$ & Val. & Test & Rand. & $\Delta$\\
\midrule
Mean training delta & +0.16 & -0.13 & -0.22 & +0.09 & +0.87 & +0.10 & +0.13 & -0.04 & +0.00 & +0.00 & -0.00 & +0.00\\
Inverse probe & +0.27 & +0.19 & -0.11 & +0.30 & +0.06 & -0.03 & -0.06 & +0.03 & +0.07 & -0.07 & -0.00 & -0.07\\
Normalized probe & +0.19 & +0.32 & -0.07 & +0.39 & +0.77 & +0.54 & +0.01 & +0.54 & +0.10 & +0.18 & +0.01 & +0.16\\
Difference of means & +0.08 & -0.04 & -0.04 & -0.00 & +1.09 & +0.03 & +0.23 & -0.20 & -0.01 & +0.01 & +0.00 & +0.01\\
PCA + ridge & +0.55 & -0.15 & -0.10 & -0.05 & +1.73 & +1.02 & +0.06 & +0.96 & +0.00 & +0.37 & +0.04 & +0.33\\
Uncentered carrier & +0.44 & -0.10 & -0.07 & -0.03 & +2.09 & +1.09 & +0.08 & +1.01 & +0.05 & +0.38 & -0.06 & +0.44\\
Immediate LoReFT & +0.32 & -0.12 & -0.07 & -0.06 & +5.06 & +3.22 & +0.24 & +2.98 & +0.31 & +0.30 & -0.03 & +0.33\\
Delayed LoReFT & +0.36 & -0.17 & -0.05 & -0.12 & +6.42 & +3.46 & +0.10 & +3.36 & +0.17 & +0.14 & +0.03 & +0.12\\
Nonlinear MLP & +0.02 & -0.02 & +0.04 & -0.07 & +1.60 & +0.76 & +0.06 & +0.70 & +0.0013 & +0.000055 & -0.0034 & +0.0034\\
Manifold--WM & +0.34 & -0.04 & +0.05 & -0.09 & +2.31 & +0.60 & +0.09 & +0.51 & -0.02 & +0.01 & -0.02 & +0.03\\
Concept DAS--WM & +0.30 & +0.05 & -0.06 & +0.10 & +0.77 & +0.44 & +0.04 & +0.39 & +0.04 & +0.31 & +0.03 & +0.28\\
\bottomrule
\end{tabular}
\end{table}

%% file: tables/random_controls_panel_storm.tex
\begin{table}[!htbp]
\centering\scriptsize\setlength{\tabcolsep}{1.7pt}
\caption{Validation, test and norm matched random controls for STORM. $\Delta$ is the test editor effect minus its random control.}\label{tab:all-random-controls-panel-storm}
\begin{tabular}{@{}lrrrrrrrrrrrr@{}}
\toprule
\multicolumn{13}{c}{ISG (pp)}\\
& \multicolumn{4}{c}{Crafter} & \multicolumn{4}{c}{Cartpole} & \multicolumn{4}{c}{CoinRun}\\
\cmidrule(lr){2-5}\cmidrule(lr){6-9}\cmidrule(lr){10-13}
Tool & Val. & Test & Rand. & $\Delta$ & Val. & Test & Rand. & $\Delta$ & Val. & Test & Rand. & $\Delta$\\
\midrule
Mean training delta & -0.49 & -0.87 & -0.45 & -0.42 & -0.20 & +0.22 & -0.48 & +0.70 & -0.00 & -0.01 & -0.00 & -0.01\\
Inverse probe & +3.91 & +2.17 & -0.55 & +2.72 & +0.29 & -0.28 & -0.21 & -0.07 & +0.01 & +0.03 & -0.29 & +0.32\\
Normalized probe & -1.24 & +0.62 & -0.09 & +0.71 & -0.09 & -0.03 & -0.13 & +0.10 & +0.04 & +0.71 & -0.17 & +0.88\\
Difference of means & -0.40 & +0.24 & +0.09 & +0.14 & +0.30 & -0.13 & -0.11 & -0.02 & +0.07 & +0.05 & +0.01 & +0.04\\
PCA + ridge & +0.48 & -0.33 & -0.53 & +0.20 & -0.23 & +0.07 & +0.07 & +0.00 & -0.02 & -0.00 & -0.00 & -0.00\\
Uncentered carrier & -0.68 & -0.17 & -0.32 & +0.15 & -0.23 & +0.00 & -0.21 & +0.21 & +0.00 & -0.00 & -0.00 & +0.00\\
Immediate LoReFT & -0.66 & -1.27 & -0.55 & -0.71 & -2.04 & +0.82 & -1.08 & +1.90 & +0.07 & +0.03 & +0.01 & +0.01\\
Delayed LoReFT & -2.24 & -3.73 & -1.25 & -2.48 & -5.06 & -1.21 & -0.66 & -0.54 & -1.04 & -1.51 & -0.18 & -1.33\\
Nonlinear MLP & -0.54 & -0.66 & -0.32 & -0.34 & -0.42 & +0.85 & -0.47 & +1.32 & 0.00 & +0.01 & +0.0011 & +0.0042\\
Manifold--WM & -0.17 & +0.13 & -0.0026 & +0.14 & +1.03 & -2.78 & +0.02 & -2.80 & +0.65 & +0.06 & -0.04 & +0.11\\
Concept DAS--WM & -0.20 & -0.04 & -0.03 & -0.01 & -0.47 & +0.08 & -0.09 & +0.17 & -0.01 & -0.01 & -0.01 & +0.01\\
\addlinespace[2pt]
\multicolumn{13}{c}{SSG@8 (pp)}\\
& \multicolumn{4}{c}{Crafter} & \multicolumn{4}{c}{Cartpole} & \multicolumn{4}{c}{CoinRun}\\
\cmidrule(lr){2-5}\cmidrule(lr){6-9}\cmidrule(lr){10-13}
Tool & Val. & Test & Rand. & $\Delta$ & Val. & Test & Rand. & $\Delta$ & Val. & Test & Rand. & $\Delta$\\
\midrule
Mean training delta & +1.87 & -0.03 & +0.12 & -0.16 & +0.20 & +0.37 & -0.24 & +0.62 & +0.02 & -0.04 & -0.01 & -0.03\\
Inverse probe & +2.42 & +2.06 & -0.20 & +2.27 & +0.47 & +0.02 & -0.20 & +0.22 & -0.07 & -0.05 & -0.14 & +0.09\\
Normalized probe & +1.62 & +1.44 & +0.43 & +1.02 & +0.37 & +0.21 & -0.15 & +0.36 & +0.14 & +0.44 & -0.18 & +0.62\\
Difference of means & +1.25 & +0.09 & +0.05 & +0.03 & +0.04 & -0.07 & -0.09 & +0.02 & -0.01 & -0.02 & +0.02 & -0.05\\
PCA + ridge & +1.36 & +0.60 & -0.49 & +1.09 & +0.60 & +0.07 & -0.06 & +0.13 & -0.01 & -0.01 & -0.01 & -0.00\\
Uncentered carrier & +1.27 & +0.17 & -0.05 & +0.22 & +0.60 & +0.03 & -0.09 & +0.12 & +0.02 & -0.01 & -0.00 & -0.01\\
Immediate LoReFT & +1.98 & -1.07 & +0.31 & -1.38 & +0.85 & +0.59 & -0.39 & +0.98 & +0.10 & +0.08 & +0.05 & +0.03\\
Delayed LoReFT & +1.50 & -0.60 & +0.01 & -0.61 & +1.20 & +0.75 & -0.16 & +0.91 & +0.18 & -1.06 & -0.01 & -1.05\\
Nonlinear MLP & +0.92 & +0.22 & +0.04 & +0.17 & +0.63 & +0.66 & +0.07 & +0.59 & +0.02 & -0.05 & -0.01 & -0.04\\
Manifold--WM & +0.88 & +0.24 & +0.08 & +0.16 & +0.75 & -1.16 & +0.08 & -1.24 & +0.04 & +0.04 & +0.01 & +0.03\\
Concept DAS--WM & +1.19 & +0.09 & +0.16 & -0.06 & 0.00 & +0.05 & -0.04 & +0.10 & -0.06 & -0.03 & -0.02 & -0.01\\
\bottomrule
\end{tabular}
\end{table}

%% file: tables/matched_patch_comparison.tex
\begin{table}[!htbp]
\centering\footnotesize\setlength{\tabcolsep}{3pt}
\caption{SSG@8 in pp for full RAP, learned editors and RAP controls matched by norm at every edit call. Each comparison shares source, event, actions and prediction noise. Full RAP retains its native magnitude.}\label{tab:matched-patch}
\begin{tabular}{@{}llrrrrr@{}}\toprule
& & Full & \multicolumn{2}{c}{Immediate} & \multicolumn{2}{c}{Delayed}\\
WM & Task & RAP & LoReFT & Matched RAP & LoReFT & Matched RAP\\\midrule
DIAMOND & Crafter & +10.76 & +0.18 & +0.48 & +0.24 & +0.48\\
DIAMOND & Cartpole & +9.01 & +0.32 & +1.08 & +3.24 & +1.97\\
DIAMOND & CoinRun & +4.14 & +0.39 & +2.01 & -1.17 & +2.01\\
DreamerV3 & Crafter & +1.24 & -0.12 & +0.38 & -0.17 & +0.26\\
DreamerV3 & Cartpole & +0.92 & +3.22 & +1.18 & +3.46 & +1.18\\
DreamerV3 & CoinRun & +1.20 & +0.30 & +1.29 & +0.14 & +1.29\\
STORM & Crafter & +3.62 & -1.07 & +3.43 & -0.60 & +3.43\\
STORM & Cartpole & +8.90 & +0.59 & +10.02 & +0.75 & +10.02\\
STORM & CoinRun & +3.35 & +0.08 & +0.29 & -1.06 & +3.58\\
\bottomrule\end{tabular}
\end{table}

%% file: tables/matched_delayed_full_norm.tex
\begin{table}[!htbp]
\centering\scriptsize\setlength{\tabcolsep}{3pt}
\caption{Bidirectional magnitude control on the frozen confirmation cohorts. ``Delayed at RAP norm'' keeps the Delayed LoReFT direction but rescales it per event and edit call to the realized full RAP norm. SSG@8 is in pp. This complements RAP evaluated at the LoReFT norm.}
\label{tab:v4-reverse-match}
\resizebox{\linewidth}{!}{%
\begin{tabular}{@{}llrrrrr@{}}\toprule
WM & Task & Delayed LoReFT & Delayed at RAP norm & Full RAP & Delayed norm & RAP norm\\\midrule
DIAMOND & Crafter & +0.24 & -3.39 & +10.76 & 22.95 & 107.43\\
DIAMOND & Cartpole & +3.24 & +3.11 & +9.01 & 37.47 & 23.90\\
DIAMOND & CoinRun & -1.17 & +0.86 & +4.14 & 580.29 & 272.40\\
DreamerV3 & Crafter & -0.17 & -0.33 & +1.24 & 5.94 & 73.64\\
DreamerV3 & Cartpole & +3.46 & +3.20 & +0.92 & 21.88 & 18.74\\
DreamerV3 & CoinRun & +0.14 & +0.13 & +1.20 & 38.68 & 31.10\\
STORM & Crafter & -0.60 & -0.05 & +3.62 & 46.68 & 33.17\\
STORM & Cartpole & +0.75 & +0.98 & +8.90 & 20.92 & 14.63\\
STORM & CoinRun & -1.06 & -0.64 & +3.35 & 53.30 & 35.27\\
\bottomrule\end{tabular}}
\end{table}

%% file: tables/horizon_sensitivity.tex
\begin{table}[!htbp]
\centering\scriptsize
\setlength{\tabcolsep}{3pt}
\caption{The fitted candidate set includes all eleven methods (the ten baseline editors and Delayed LoReFT). Horizon sensitivity on the frozen confirmation reports. Each entry uses the scoreability threshold $\lceil0.75H\rceil$ and aggregation with equal source weights. Gain sequences follow $H=1,2,4,8$. Immediate and Delayed denote the two LoReFT objectives. RAP is shown separately and is not ranked with fitted editors.}\label{tab:horizon-sensitivity}
\resizebox{\textwidth}{!}{%
\begin{tabular}{@{}lllrlll@{}}\toprule
WM & Task & H=8 fitted leader & \multicolumn{1}{c}{Leader by $H=1,2,4,8$} & Fitted leader gains & RAP gains & Stable?\\\midrule
DIAMOND & Crafter & Mean & InvProbe/InvProbe/InvProbe/Mean & +0.60/+0.69/+0.63/+0.31 & +12.81/+12.69/+12.20/+10.76 & no\\
DIAMOND & Cartpole & Delayed & Delayed/Delayed/Delayed/Delayed & +0.91/+1.58/+2.18/+3.24 & +6.00/+6.13/+6.93/+9.01 & yes\\
DIAMOND & CoinRun & NormProbe & NormProbe/NormProbe/NormProbe/NormProbe & +4.15/+3.97/+3.36/+2.25 & +7.13/+6.71/+5.77/+4.14 & yes\\
DreamerV3 & Crafter & NormProbe & InvProbe/InvProbe/NormProbe/NormProbe & +0.08/+0.16/+0.27/+0.32 & +2.11/+1.92/+1.54/+1.24 & no\\
DreamerV3 & Cartpole & Delayed & Delayed/Delayed/Delayed/Delayed & +2.53/+2.69/+2.97/+3.46 & +0.56/+0.69/+0.87/+0.92 & yes\\
DreamerV3 & CoinRun & Carrier & PCA+ridge/PCA+ridge/PCA+ridge/Carrier & +0.22/+0.23/+0.31/+0.38 & +1.08/+1.09/+1.18/+1.20 & no\\
STORM & Crafter & InvProbe & InvProbe/InvProbe/InvProbe/InvProbe & +1.38/+1.95/+2.59/+2.06 & +5.21/+4.91/+4.57/+3.62 & yes\\
STORM & Cartpole & Delayed & Immediate/Immediate/Nonlinear MLP/Delayed & +0.70/+0.64/+0.64/+0.75 & +8.52/+8.79/+8.94/+8.90 & no\\
STORM & CoinRun & NormProbe & NormProbe/NormProbe/NormProbe/NormProbe & +0.65/+0.64/+0.56/+0.44 & +4.58/+4.26/+3.81/+3.35 & yes\\
\bottomrule\end{tabular}}
\end{table}

%% file: tables/action_sensitivity.tex
\begin{table}[!htbp]
\centering\scriptsize
\setlength{\tabcolsep}{3pt}
\caption{The fitted candidate set includes all eleven methods (the ten baseline editors and Delayed LoReFT). Sensitivity to action sequences on the same frozen events. Cartpole and Crafter are split at the task median of mean action variation after editing; CoinRun is split by whether its schedule of held actions over eight steps switches in the measured window. $L/H$ denotes lower or higher variation, and constant or switched actions for CoinRun. Immediate and Delayed denote the two LoReFT objectives.}\label{tab:action-sensitivity}
\resizebox{\textwidth}{!}{%
\begin{tabular}{@{}lllrrrr@{}}\toprule
WM & Task & Split & $n_L/n_H$ sources & Fitted leader $L/H$ & RAP $L/H$ & Same fitted leader?\\\midrule
DIAMOND & Crafter & lower / higher action variation & 22/18 & InvProbe +0.54/Mean +0.54 & +10.96/+10.31 & no\\
DIAMOND & Cartpole & lower / higher action variation & 39/39 & Delayed +2.69/Delayed +3.85 & +10.36/+8.48 & yes\\
DIAMOND & CoinRun & constant / switched & 19/44 & NormProbe +0.78/NormProbe +2.68 & +2.55/+4.41 & yes\\
DreamerV3 & Crafter & lower / higher action variation & 22/18 & NormProbe +0.43/InvProbe +0.18 & +0.63/+1.99 & no\\
DreamerV3 & Cartpole & lower / higher action variation & 39/39 & Delayed +3.36/Delayed +3.73 & +0.82/+1.09 & yes\\
DreamerV3 & CoinRun & constant / switched & 19/44 & NormProbe +0.13/PCA+ridge +0.49 & +0.95/+1.26 & no\\
STORM & Crafter & lower / higher action variation & 22/18 & NormProbe +1.35/InvProbe +3.38 & +2.24/+4.81 & no\\
STORM & Cartpole & lower / higher action variation & 39/39 & Delayed +1.14/Nonlinear MLP +0.71 & +10.22/+7.43 & no\\
STORM & CoinRun & constant / switched & 19/44 & NormProbe +0.24/NormProbe +0.56 & +3.02/+3.44 & yes\\
\bottomrule\end{tabular}}
\end{table}

%% file: tables/noise_sensitivity.tex
\begin{table}[!htbp]
\centering\scriptsize
\setlength{\tabcolsep}{3pt}
\caption{The fitted candidate set includes all eleven methods (the ten baseline editors and Delayed LoReFT). Sensitivity to prediction noise. Columns list the three preregistered seeds (69000/69001/69002); each seed is aggregated independently over events and sources. Immediate and Delayed denote the two LoReFT objectives. The seeds are robustness strata, not additional independent datasets.}\label{tab:noise-sensitivity}
\begin{tabular}{@{}lllll@{}}\toprule
WM & Task & Fitted winner by seed & Fitted SSG@8 by seed & RAP SSG@8 by seed\\\midrule
DIAMOND & Crafter & Mean/Mean/Delayed & +0.32/+0.32/+0.30 & +10.90/+10.54/+10.84\\
DIAMOND & Cartpole & Delayed/Delayed/Delayed & +3.16/+3.30/+3.27 & +9.07/+9.01/+8.94\\
DIAMOND & CoinRun & NormProbe/NormProbe/NormProbe & +2.08/+2.48/+2.20 & +4.04/+4.31/+4.07\\
DreamerV3 & Crafter & InvProbe/PCA+ridge/NormProbe & +0.45/+0.03/+0.71 & +1.07/+1.43/+1.21\\
DreamerV3 & Cartpole & Delayed/Delayed/Delayed & +3.93/+3.78/+2.67 & +0.76/+1.18/+0.84\\
DreamerV3 & CoinRun & Carrier/PCA+ridge/Immediate & +0.39/+0.51/+0.27 & +1.37/+1.21/+1.02\\
STORM & Crafter & InvProbe/InvProbe/InvProbe & +1.74/+2.29/+2.16 & +3.78/+4.38/+2.70\\
STORM & Cartpole & NormProbe/Nonlinear MLP/Delayed & +0.51/+1.56/+1.46 & +8.94/+9.56/+8.20\\
STORM & CoinRun & NormProbe/NormProbe/NormProbe & +0.27/+0.38/+0.66 & +3.19/+3.55/+3.30\\
\bottomrule\end{tabular}
\end{table}

%% file: figures/temporal_regimes_figure.tex
\begin{figure}[!htbp]
\centering\includegraphics[width=\linewidth]{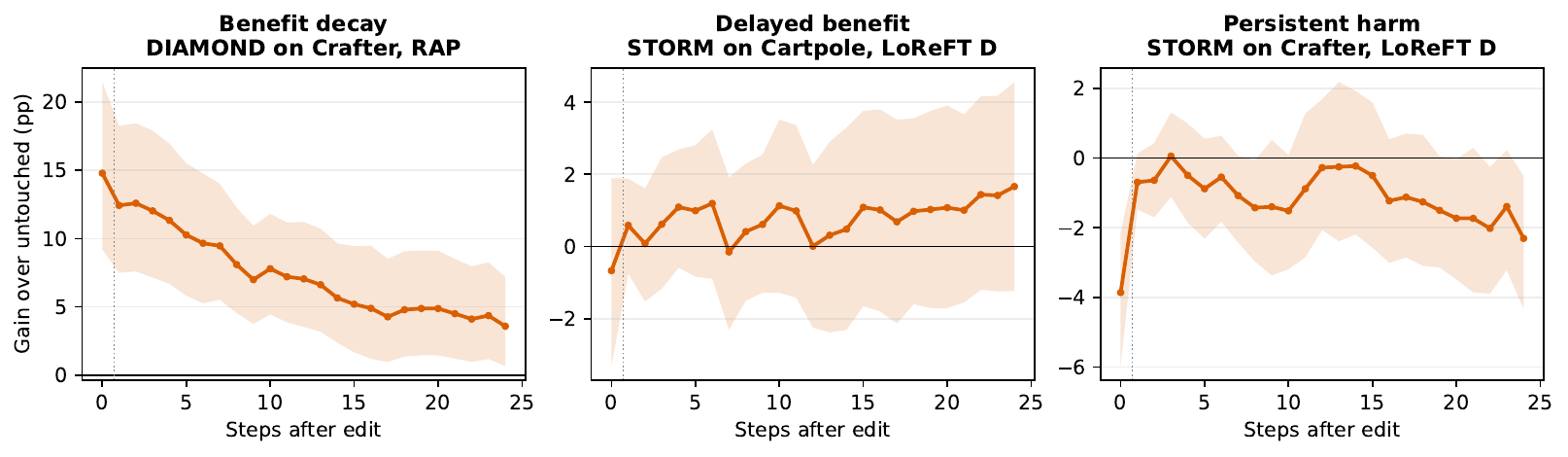}
\caption{\textbf{Representative effects over 24 prediction steps.} Source weighted means and 95\% bootstrap intervals illustrate three recurring regimes: a large RAP benefit that decays, a fitted effect that emerges after the edited prediction, and harm that persists during autonomous prediction.}\label{fig:temporal-regimes}
\end{figure}

%% file: tables/restoration_all.tex
\begin{longtable}{@{}lllrrrr@{}}
\caption{Component restoration results across all models. Original and restored gains average steps 9 to 24; losses are original minus restored gain. Positive loss means restoration removes benefit, while negative loss means restoration improves the edited rollout. Values are means in pp with equal source weights.}\label{tab:v4-restoration-all}\\
\toprule WM & Task & Edit & Original & Restore latent & Restore memory & $L_z/L_m$\\\midrule\endfirsthead
\toprule WM & Task & Edit & Original & Restore latent & Restore memory & $L_z/L_m$\\\midrule\endhead
DIAMOND & Crafter & RAP & +5.43 & +0.01 & +5.40 & +5.42/+0.02\\
DIAMOND & Crafter & LoReFT D & +1.31 & -0.02 & +1.26 & +1.33/+0.05\\
DIAMOND & Crafter & LoReFT I & +0.43 & +0.03 & +0.43 & +0.39/-0.01\\
DIAMOND & Cartpole & RAP & +12.88 & -1.20 & +11.97 & +14.09/+0.91\\
DIAMOND & Cartpole & LoReFT D & +3.32 & -1.61 & +4.08 & +4.93/-0.76\\
DIAMOND & Cartpole & LoReFT I & +0.47 & -0.08 & +0.49 & +0.55/-0.02\\
DIAMOND & CoinRun & RAP & +1.78 & -1.24 & +1.56 & +3.02/+0.22\\
DIAMOND & CoinRun & LoReFT D & +0.41 & -0.96 & -1.85 & +1.37/+2.26\\
DIAMOND & CoinRun & LoReFT I & +2.59 & +0.39 & +2.65 & +2.20/-0.06\\
DreamerV3 & Crafter & RAP & +0.60 & +0.47 & +0.17 & +0.13/+0.43\\
DreamerV3 & Crafter & LoReFT D & -0.10 & -0.18 & +0.05 & +0.08/-0.15\\
DreamerV3 & Crafter & LoReFT I & +0.02 & -0.16 & -0.05 & +0.18/+0.07\\
DreamerV3 & Cartpole & RAP & +1.22 & +1.09 & +0.12 & +0.14/+1.10\\
DreamerV3 & Cartpole & LoReFT D & +4.95 & +4.77 & +0.28 & +0.18/+4.68\\
DreamerV3 & Cartpole & LoReFT I & +4.75 & +4.74 & +0.20 & +0.01/+4.55\\
DreamerV3 & CoinRun & RAP & +0.61 & +0.62 & +0.04 & -0.01/+0.57\\
DreamerV3 & CoinRun & LoReFT D & +0.57 & +0.49 & -0.01 & +0.08/+0.59\\
DreamerV3 & CoinRun & LoReFT I & +0.17 & +0.30 & -0.06 & -0.13/+0.23\\
STORM & Crafter & RAP & +1.88 & +0.87 & +1.39 & +1.01/+0.49\\
STORM & Crafter & LoReFT D & -1.21 & +0.11 & -0.28 & -1.32/-0.93\\
STORM & Crafter & LoReFT I & -1.44 & -0.13 & -0.21 & -1.32/-1.23\\
STORM & Cartpole & RAP & +5.54 & +2.44 & +2.74 & +3.10/+2.80\\
STORM & Cartpole & LoReFT D & +0.93 & +1.21 & +0.40 & -0.28/+0.53\\
STORM & Cartpole & LoReFT I & +1.13 & +0.95 & +0.45 & +0.18/+0.69\\
STORM & CoinRun & RAP & +0.61 & -0.10 & +0.25 & +0.70/+0.36\\
STORM & CoinRun & LoReFT D & -0.53 & -0.07 & -0.77 & -0.47/+0.24\\
STORM & CoinRun & LoReFT I & +0.09 & -0.14 & -0.05 & +0.23/+0.14\\
\bottomrule\end{longtable}

%% file: figures/pathway_diamond.tex
\begin{figure}[p]
\centering\includegraphics[width=\linewidth]{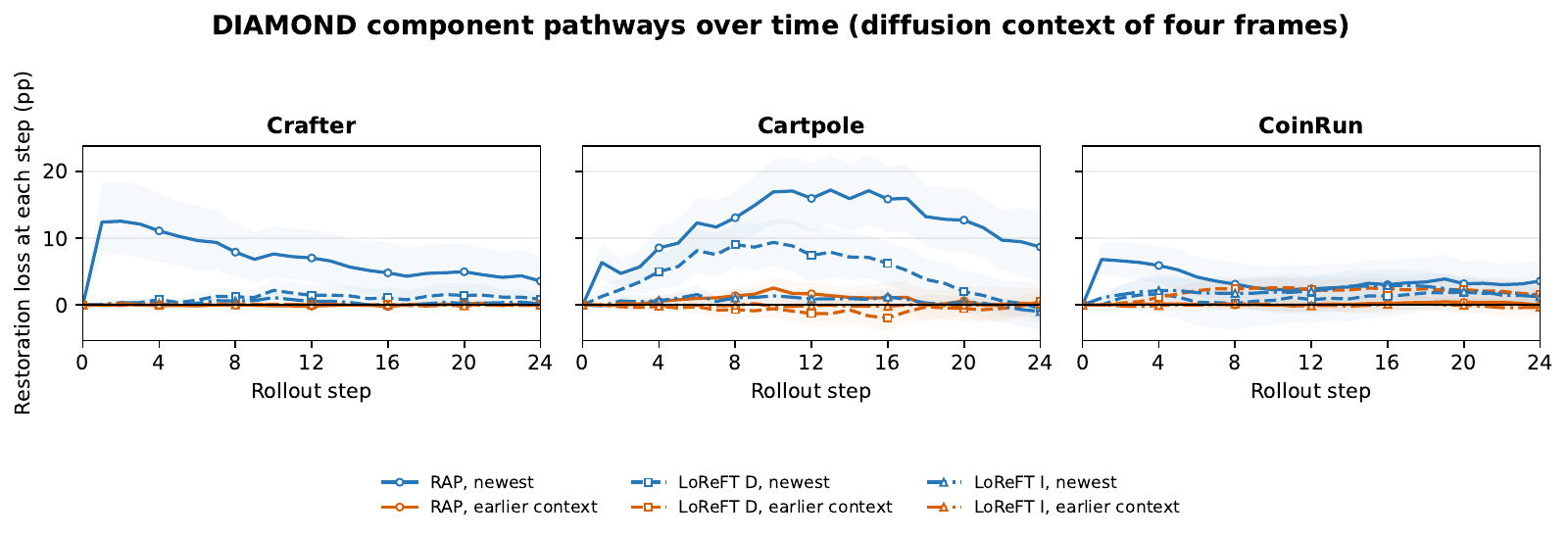}
\caption{\textbf{DIAMOND component pathways over 24 prediction steps.} Restoration loss is shown for the newest generated frame and the three earlier context frames under RAP, Immediate LoReFT and Delayed LoReFT. Shading gives 95\% source bootstrap intervals.}\label{fig:pathway-diamond}
\end{figure}

%% file: figures/pathway_dreamer.tex
\begin{figure}[p]
\centering\includegraphics[width=\linewidth]{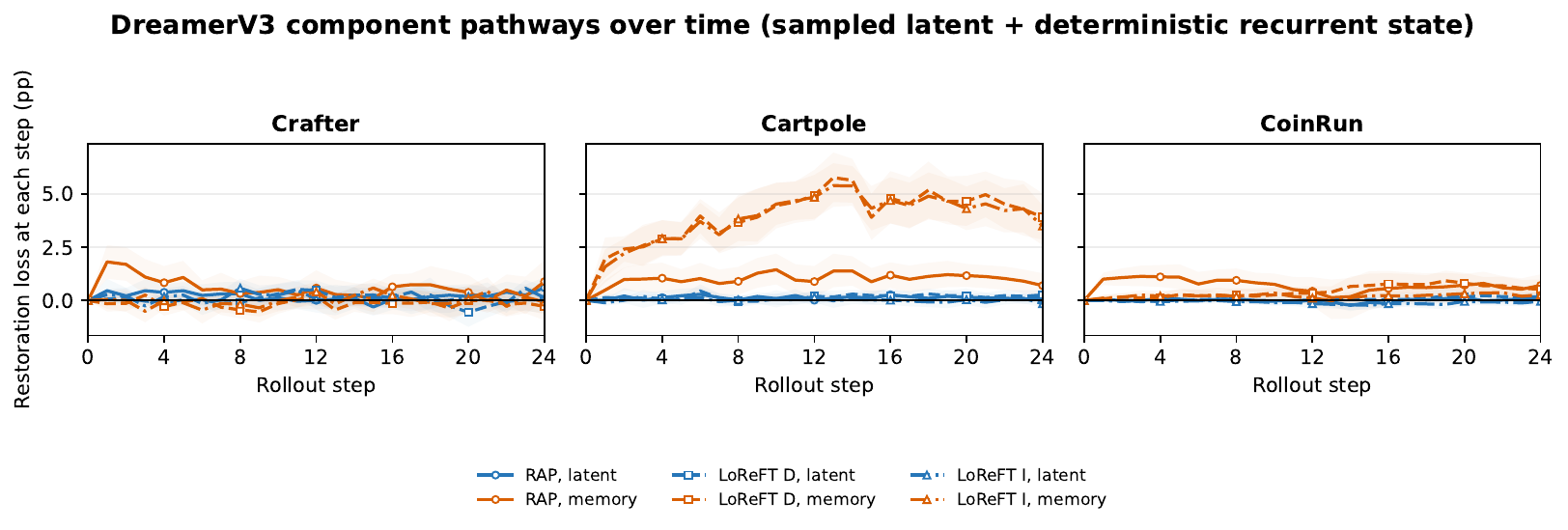}
\caption{\textbf{DreamerV3 component pathways over 24 prediction steps.} Restoration loss is shown for the sampled latent and deterministic recurrent memory under RAP, Immediate LoReFT and Delayed LoReFT. Shading gives 95\% source bootstrap intervals.}\label{fig:pathway-dreamer}
\end{figure}

%% file: figures/pathway_storm.tex
\begin{figure}[p]
\centering\includegraphics[width=\linewidth]{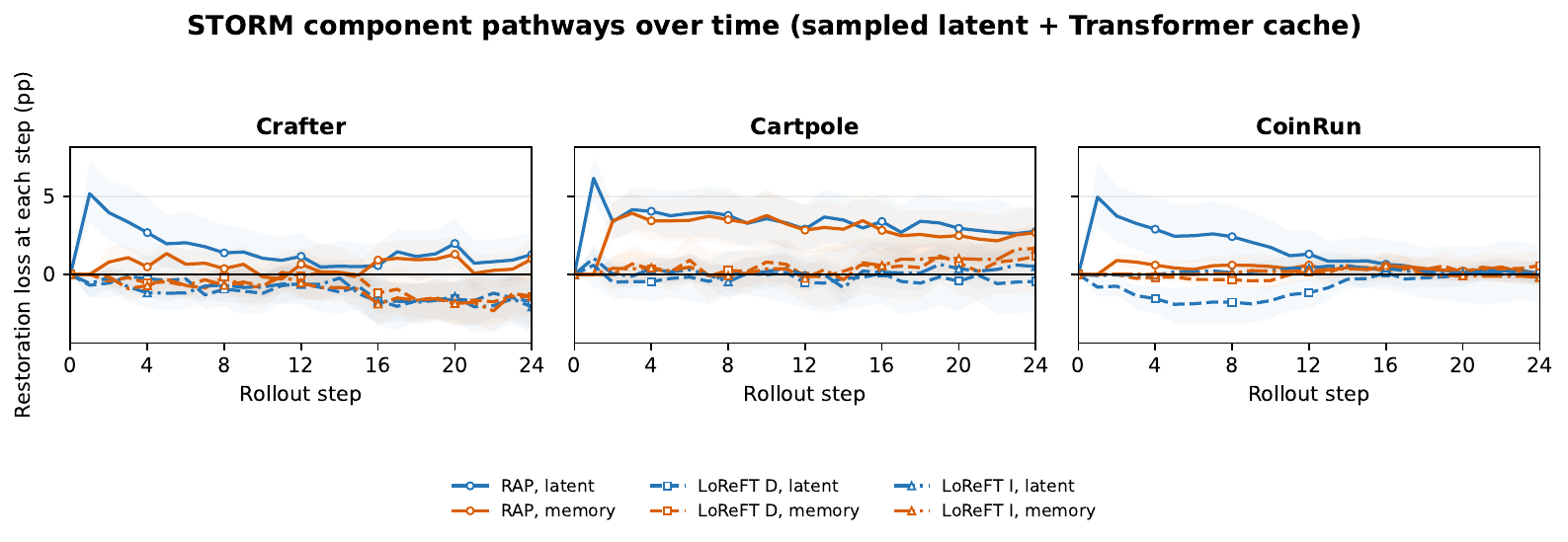}
\caption{\textbf{STORM component pathways over 24 prediction steps.} Restoration loss is shown for the sampled latent and transformer cache under RAP, Immediate LoReFT and Delayed LoReFT. Shading gives 95\% source bootstrap intervals.}\label{fig:pathway-storm}
\end{figure}

%% file: tables/pathway_statistics.tex
\begin{table}[!htbp]
\centering\scriptsize\setlength{\tabcolsep}{4pt}
\caption{Later gain and restoration loss over steps 9 to 24, in pp. Loss is edited gain minus gain after restoring the named component at step 1. Brackets show 95\% source bootstrap intervals.}\label{tab:pathway-loss}
\begin{tabular}{@{}llrrrr@{}}\toprule
Cohort & Edit & Sources & Original & Latent loss & Memory loss\\\midrule
DreamerV3 Cartpole & RAP & 48 & \shortstack{$+1.22$\\$[+0.74,+1.70]$} & \shortstack{$+0.14$\\$[+0.02,+0.28]$} & \shortstack{$+1.10$\\$[+0.66,+1.54]$}\\
DreamerV3 Cartpole & Delayed LoReFT & 48 & \shortstack{$+4.95$\\$[+4.10,+5.86]$} & \shortstack{$+0.18$\\$[+0.02,+0.33]$} & \shortstack{$+4.68$\\$[+3.80,+5.58]$}\\
STORM Cartpole & RAP & 48 & \shortstack{$+5.54$\\$[+3.77,+7.34]$} & \shortstack{$+3.10$\\$[+1.70,+4.56]$} & \shortstack{$+2.80$\\$[+1.42,+4.26]$}\\
STORM Cartpole & Delayed LoReFT & 48 & \shortstack{$+0.93$\\$[-1.65,+3.45]$} & \shortstack{$-0.28$\\$[-2.03,+1.59]$} & \shortstack{$+0.53$\\$[-1.25,+2.15]$}\\
STORM Crafter & RAP & 37 & \shortstack{$+1.88$\\$[+0.44,+3.50]$} & \shortstack{$+1.01$\\$[-0.17,+2.33]$} & \shortstack{$+0.49$\\$[-0.31,+1.26]$}\\
STORM Crafter & Delayed LoReFT & 37 & \shortstack{$-1.21$\\$[-2.78,+0.29]$} & \shortstack{$-1.32$\\$[-2.49,-0.34]$} & \shortstack{$-0.93$\\$[-2.02,+0.07]$}\\
\bottomrule\end{tabular}
\end{table}

%% file: tables/random_restoration_controls.tex
\begin{table}[!htbp]
\centering\scriptsize
\setlength{\tabcolsep}{3pt}
\caption{Matched random controls for hybrid state restoration, in pp. The final column subtracts the equal norm random restoration loss from the semantic restoration loss; brackets show paired 95\% source bootstrap intervals.}\label{tab:random-restoration-controls}
\resizebox{\textwidth}{!}{%
\begin{tabular}{@{}llllrrr@{}}\toprule
WM & Task & Edit & Restored & Random loss & Semantic loss & Semantic minus random loss (95\% CI)\\\midrule
DreamerV3 & Cartpole & RAP & Latent & +0.09 & +0.14 & +0.05 [-0.11, +0.22]\\
DreamerV3 & Cartpole & RAP & Memory & +0.15 & +1.10 & +0.96 [+0.41, +1.49]\\
DreamerV3 & Cartpole & Delayed LoReFT & Latent & -0.04 & +0.18 & +0.22 [+0.04, +0.39]\\
DreamerV3 & Cartpole & Delayed LoReFT & Memory & +0.05 & +4.68 & +4.62 [+3.78, +5.46]\\
STORM & Cartpole & RAP & Latent & -0.45 & +3.10 & +3.55 [+2.00, +5.14]\\
STORM & Cartpole & RAP & Memory & -0.04 & +2.80 & +2.84 [+1.41, +4.38]\\
STORM & Cartpole & Delayed LoReFT & Latent & -0.90 & -0.28 & +0.62 [-1.35, +2.72]\\
STORM & Cartpole & Delayed LoReFT & Memory & -0.35 & +0.53 & +0.88 [-0.95, +2.67]\\
STORM & Crafter & RAP & Latent & -0.14 & +1.01 & +1.15 [-0.22, +2.62]\\
STORM & Crafter & RAP & Memory & -0.05 & +0.49 & +0.54 [-0.36, +1.45]\\
STORM & Crafter & Delayed LoReFT & Latent & -0.31 & -1.32 & -1.01 [-2.22, +0.10]\\
STORM & Crafter & Delayed LoReFT & Memory & +0.19 & -0.93 & -1.12 [-2.82, +0.34]\\
\bottomrule\end{tabular}}
\end{table}